\documentclass[5p]{elsarticle}

\usepackage{subcaption}
\usepackage{dirtytalk}
\usepackage{amssymb,amsmath}
\usepackage{bm}
\usepackage[export]{adjustbox}
\usepackage{xcolor}
\usepackage{flushend}
\usepackage{stfloats}

\definecolor{our-baby}{HTML}{C3FDFC}
\definecolor{our-bg}{HTML}{EEF7E9}
\definecolor{our-blue}{rgb}{0.367, 0.504, 0.707}
\definecolor{our-gray}{gray}{0.96}
\definecolor{our-green}{HTML}{6CFFCE}
\definecolor{our-ink}{HTML}{000000}
\definecolor{our-layer}{HTML}{D4EEF2}
\definecolor{our-light-ink}{HTML}{5D7073}
\definecolor{our-light-yellow}{HTML}{EDF1BF}
\definecolor{our-pink}{HTML}{F9B2B3}
\definecolor{our-red}{rgb}{0.923, 0.386, 0.209}
\definecolor{our-sta}{HTML}{BFEFFF}
\definecolor{our-yellow}{HTML}{FFE699}

\definecolor{NeoRed}{RGB}{0, 255, 255}
\definecolor{EpiOrg}{RGB}{255, 0, 255}
\definecolor{InfGrn}{RGB}{0, 255, 0}
\definecolor{ConBlu}{RGB}{255, 255, 0}
\definecolor{DeadYel}{RGB}{61, 100, 255}

\graphicspath{{fig/}}

\usepackage{booktabs}
\usepackage{multirow}
\usepackage{multicol}
\usepackage{colortbl}
\newcolumntype{g}{>{\columncolor{our-gray}}c} 
\newcommand{\wb}{\cellcolor{white}} 

\usepackage[
  breaklinks=true,
  bookmarks=false,
  colorlinks=true,
]{hyperref}

\usepackage[capitalize]{cleveref}
\crefname{section}{Sec.}{Sec.}
\crefname{table}{Tab.}{Tab.}
\crefname{figure}{Fig.}{Fig.}

\usepackage{etoolbox}
\usepackage{orcidlink}

\usepackage{tikz}
\usetikzlibrary{
  matrix,
  arrows.meta,
  backgrounds,
  fit,
  positioning,
  shapes.geometric,
  calc
}

\makeatletter
\let\@afterindenttrue\@afterindentfalse
\newcommand{\authororcid}[1]{\g@addto@macro\elsauthors{\nobreak\,\orcidlink{#1}}}
\makeatother

\biboptions{sort&compress}

\journal{Computer Methods and Programs in Biomedicine}

\newpageafter{author}

\begin{document}

\hypersetup{linkcolor=our-blue, urlcolor=our-blue, citecolor=our-red}

\begin{frontmatter}

\title{LSP-DETR: Efficient and Scalable\texorpdfstring{\\}{ }Nuclei Segmentation in Whole-Slide Images}

\author[fi]{Matěj Pekár}\authororcid{0009-0008-6687-5199}
\author[fi]{Vít Musil\corref{cor}\fnref{phone}}\authororcid{0000-0001-6083-227X}
\ead{musil@fi.muni.cz}
\author[mou]{Rudolf Nenutil}\authororcid{0000-0003-0068-9760}
\author[bbmri]{Petr Holub}\authororcid{0000-0002-5358-616X}
\author[fi]{Tomáš Brázdil}\authororcid{0000-0002-4547-3261}

\affiliation[fi]{
    organization={Masaryk~University, Faculty~of~Informatics},
    addressline={Botanická~68a},
    city={Brno},
    postcode={602~00},
    country={Czech~Republic}
}
\affiliation[bbmri]{
    organization={BBMRI-ERIC},
    addressline={Neue Stiftingtalstraße 2/B/6},
    city={Graz},
    postcode={8010},
    country={Austria}
}
\affiliation[mou]{
    organization={Masaryk Memorial Cancer Institute},
    addressline={Žlutý kopec~7},
    city={Brno},
    postcode={656~53},
    country={Czech Republic}
}
\cortext[cor]{Corresponding author}

\begin{abstract}
\paragraph{Background and Objective}
Precise and scalable instance segmentation of cell nuclei is a fundamental prerequisite for computational pathology, yet gigapixel whole-slide images (WSIs) pose significant computational challenges. While patch-based processing is standard during training, existing methods are often limited to small tile sizes during inference due to architectural bottlenecks or reliance on computationally expensive post-processing steps for instance separation. 
We introduce a faster, scalable, and end-to-end framework capable of processing large-scale image tiles while accurately modeling biologically realistic overlapping nuclei.

\paragraph{Methods}
We propose LSP-DETR (Local Star Polygon DEtection TRansformer).
The model represents nuclei as star-convex polygons and employs a lightweight transformer with linear complexity, enabling the processing of high-resolution images in a single forward pass. A novel radial distance loss accommodates annotation uncertainty, allowing the segmentation of overlapping nuclei to emerge naturally without explicit overlap labels.

\paragraph{Results}
LSP-DETR achieves state-of-the-art efficiency, with an inference time of 0.45 s/mm\textsuperscript{2}, representing a 3.2\texttimes{} speedup over StarDist, the next-fastest method. On PanNuke, the model achieves competitive accuracy (67.5 bPQ), while yielding an F$_1$-score of 0.964 in polygon overlap when evaluated against consensus annotations from two expert pathologists. Furthermore, it outperforms larger models such as LKCell in terms of generalization robustness, reaching an F$_1$-score of 85.0 on MoNuSeg.

\paragraph{Conclusions}
LSP-DETR bridges the gap between high-fidelity segmentation and practical clinical requirements by eliminating heuristic post-processing.
By providing a scalable, linear-complexity solution that naturally handles overlaps between nuclei, this framework sets a new direction for efficient high-throughput WSI analysis in digital pathology.
\end{abstract}


\end{frontmatter}

\section{Introduction}
Assessment of cell nuclear morphology and tissue architecture is central to histopathological diagnosis, particularly in oncology. Abnormalities in nuclear size and shape, chromatin distribution, and nucleolar prominence are established cytological criteria of malignancy \cite{zink2004nuclear,fischer2010cytologic}. In several cancers, nuclear pleomorphism and mitotic activity are incorporated into histological grading systems with prognostic value; the Nottingham grading system for invasive breast carcinoma is a canonical example \cite{elston1991pathological}. At the tissue scale, nuclear density and spatial organization contribute to the recognition of glandular architecture and the arrangement of tumor, stromal, and immune compartments \cite{gurcan2009histopathological,heindl2015mapping}. In digital pathology, nuclei instance segmentation converts these visual characteristics into per-cell boundaries, morphometric measurements, and spatial coordinates.

Significant advances have been made in nuclei segmentation algorithms in digital pathology and cell biology over the last decade~\cite{magou2024overview}, leading to their integration into widely used software platforms such as QuPath and HALO.
These algorithms are an essential prerequisite for more sophisticated artificial intelligence applications, including cell-based graph neural networks, which analyze the intricate relationships and spatial organization of nuclei (see~\cite{brusse2025GNN} for a comprehensive overview).

\begin{figure*}[t!]
  \centering
  \begin{subfigure}
    [t]{\columnwidth}
    \centering
    \includegraphics[width=\linewidth]{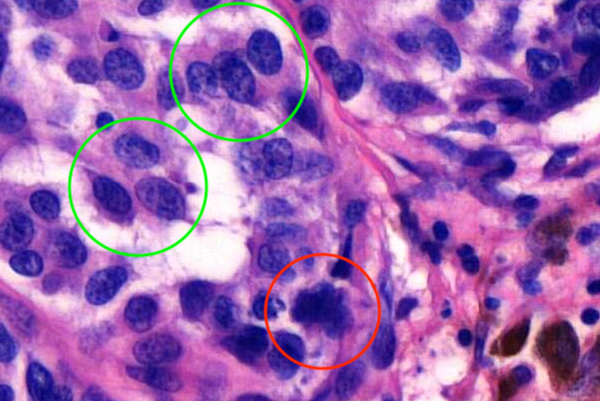}
    \caption{
      An example of an H\&E-stained image.
      The green areas highlight delineated or overlapping nuclei.
      In the red area, accurate nuclei segmentation is impossible.
    }
    \label{fig:nuclei-background:regions}
  \end{subfigure}
  \hfill
  \begin{subfigure}
    [t]{\columnwidth}
    \centering
    \includegraphics[width=\linewidth]{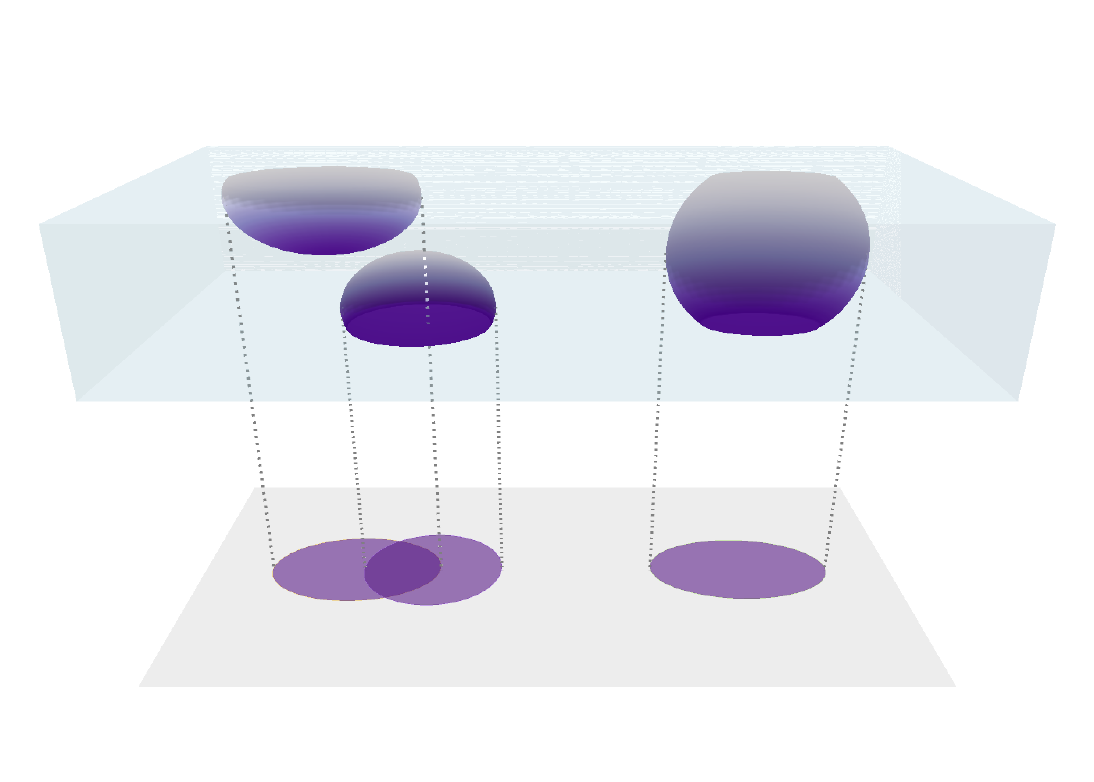}
    \caption{Illustration of how nuclei overlap as a result of tissue slicing and scanning.
    }
    \label{fig:nuclei-background:overlaps}
  \end{subfigure}%
  \caption{Ambiguous boundaries and overlaps pose one of the challenges in
  nuclei segmentation.}
  \label{fig:nuclei-background}
\end{figure*}

\subsection{Problem Formulation}
\label{sec:prob_form}

We aim to design a \emph{fast, scalable, and end-to-end trainable} algorithm for \emph{cell nuclei instance segmentation in whole-slide images (WSIs) of hematoxylin-eosin-stained tissue samples}, including \emph{overlapping nuclei} in two-dimensional WSIs due to their configuration in three-dimensional physical tissue (see Fig.~\ref{fig:nuclei-background:overlaps}).

Our goal is to develop a universally applicable system across a wide range of WSIs from different tissues, organs, and pathological alterations.
We concentrate on hematoxylin-eosin-stained WSIs, but our method can also be applied to other types of staining, assuming detailed nuclei annotations.
Solving the cell nuclei instance segmentation problem in WSIs poses several key challenges:

\paragraph{Boundary Ambiguity and Physical Overlaps}
Due to the three-dimensional physical organization of tissue sections, spherical cell nuclei can overlap or occlude one another when projected onto a two-dimensional digital WSI scan, as illustrated in~\cref{fig:nuclei-background:overlaps}. The frequency of these overlaps depends on the tissue type---for example, they are relatively uncommon in stroma but relatively common in areas with high cell density.
Standard digital pathology benchmarks typically enforce non-overlapping boundaries during manual annotation due to software constraints.
Furthermore, precisely delineating true biological cell occlusions is highly labor-intensive and time-consuming for expert pathologists. Consequently, traditional frameworks face a severe data paradox: they are forced to learn rigid, artificial boundaries that conflict with the underlying biological reality. On the other hand, because fully annotated three-dimensional data are difficult to obtain, there is a clear need for methods that can leverage existing datasets with incomplete overlap annotations while still inferring true nuclei shapes.
The same mismatch also creates an evaluation paradox: standard pixel-overlap metrics penalize biologically plausible overlapping predictions for crossing the artificial boundaries imposed by non-overlapping ground~truth.

\paragraph{Large Data Variability}
Whole-slide images can vary significantly across multiple dimensions. First, tissue morphology differs substantially across organs, pathological alterations, and even within a single WSI. Nuclei exhibit wide variability in shape, size, and density, and their boundaries are often ambiguous due to overlapping structures and staining limitations, as illustrated in~\cref{fig:nuclei-background:regions}. Second, a generic and universally applicable algorithm must account for variations in tissue preparation, staining protocols, and pathology practices that differ across laboratories.

\paragraph{Large Image Size}
Gigapixel WSIs cannot be held in memory at full resolution, so they must be divided for processing. In many existing pipelines, the usable tile size is limited by factors other than memory. Some vision-transformer architectures are tied to a fixed input size because of their positional encoding or attention design, and therefore cannot adapt the tile size to the available hardware. Other architectures, including fully convolutional models, can accept larger tiles but commonly rely on computationally expensive post-processing to form instances. The cost of this post-processing grows with tile size, making large-tile inference inefficient. Thus, tiling remains a practical requirement either because the model cannot accommodate larger inputs or because the associated post-processing becomes prohibitively costly. Small tiles introduce two additional inefficiencies. First, avoiding cut-off nuclei requires overlapping margins between adjacent tiles, and the smaller the tiles, the larger this margin grows relative to their content, inflating redundant computation. Second, a nucleus split across two tiles yields conflicting predictions that must be reconciled by custom heuristic stitching routines, which frequently split or duplicate cells. An ideal method would remove the architectural constraint entirely, leaving memory as the sole limit on tile size, so that tiles can be made as large as the hardware allows, and, in the limit of sufficient memory, dispensed with altogether.

\bigskip
These characteristics of the nuclei segmentation problem pose difficulties across all phases of the development pipeline, from the acquisition of large-scale WSIs through the robustness and generalizability of segmentation algorithms to evaluation, in which generating high-quality annotations of individual cell nuclei is both labor-intensive and time-consuming.

\subsection{State of the Art}

The cell nuclei segmentation problem has been extensively studied in various contexts~\cite{graham2019hover, 11079405}.
Classical approaches, typically based on watershed methods, are still employed in small-scale applications.
These methods are simple, interpretable, and computationally efficient.
However, they struggle to cope with the data variability encountered in large-scale applications involving numerous images from diverse sources.

Deep-learning methods improve robustness across heterogeneous tissue appearances, but existing approaches retain important limitations for WSI-scale inference.
Some are tied to specific image sizes~\cite{Huang_2023_ICCV,CellViT,obeid2022nucdetr, pina2024celldetr}, leading to inefficiencies in training or inference, as the annotated data usually consist of small tiles.
Others rely on computationally expensive post-processing pipelines~\cite{chen2023cpp,LKCell,graham2019hover,StarDist,StarDistClassification}, which hinder the accurate segmentation of overlapping nuclei and add unnecessary complexity.

\subsection{Contribution: LSP-DETR}
To bridge the gap between pixel-level optimization and instance-level biological reality, we introduce the Local Star Polygon DEtection TRansformer (LSP-DETR). The model applies a lightweight, linear-complexity transformer to localized grid queries to predict a set of nuclei represented as star-convex polygons. Our contributions span the representation, architecture, learning objective, evaluation protocol, and cross-dataset validation and are summarized in \cref{tab:comparison}.

\begin{table}[tb]
  \centering
  \footnotesize
  \caption{
    Comparison of nuclei segmentation methods.
    The summarized characteristics include post-processing requirements, overlap-prediction capabilities, adaptability to arbitrary input sizes during inference, and inference speed.
    }
  \begingroup
  \setlength{\tabcolsep}{1ex}
  \begin{tabular*}{\linewidth}{@{\ \extracolsep{\fill}}l@{\!}cccr}
    \toprule
    Method
        & End-to-end & Overlaps    & Any Size & Inference Time  \\
    \midrule
    StarDist~\cite{StarDist}
        & --          & \checkmark* & \checkmark     &  1.42 s/mm\textsuperscript{2}\phantom{\textsuperscript{\dag}} \\
    CPP-Net~\cite{chen2023cpp}
        & --          & \checkmark* & \checkmark     & 21.21 s/mm\textsuperscript{2}\phantom{\textsuperscript{\dag}}\\
    HoVer-NeXt~\cite{baumann2024hover}
        & --          & --           & \checkmark     &  2.38 s/mm\textsuperscript{2}\textsuperscript{\dag}   \\
    LKCell~\cite{LKCell}
        & --          & --           & \checkmark     &  4.39 s/mm\textsuperscript{2}\phantom{\textsuperscript{\dag}}\\
    CellViT~\cite{CellViT}
        & --          & --           & --             & 13.24 s/mm\textsuperscript{2}\textsuperscript{\ddag} \\
    \textbf{LSP-DETR}
        & \checkmark & \checkmark  & \checkmark     &  0.45 s/mm\textsuperscript{2}\phantom{\textsuperscript{\dag}} \\
    \bottomrule\\[-2ex]
    \multicolumn{5}{p{0.95\linewidth}}{\footnotesize \textsuperscript{*}~Method was penalized for segmenting overlaps during training.} \\
    \multicolumn{5}{p{0.95\linewidth}}{\footnotesize \textsuperscript{\dag}~Inference time excludes post-processing.} \\
    \multicolumn{5}{p{0.95\linewidth}}{\footnotesize \textsuperscript{\ddag}~CellViT: literature-derived A40 timing not directly comparable to the H100 results.} \\
  \end{tabular*}
  \endgroup
  \label{tab:comparison}
  \vspace{1ex}
\end{table}

\paragraph{End-to-End Direct Set Prediction}
Nuclei segmentation is naturally a set-prediction problem: an image contains an unordered set of instances whose number varies across fields of view. DETR \cite{detr} provides a direct formulation of precisely this structure through bipartite matching and parallel instance prediction, rather than approximating it with pixel grouping, watershed, or NMS. Accordingly, LSP-DETR receives an H\&E-stained image as input and directly outputs the positions and boundaries of the nuclei, structurally eliminating non-differentiable external components that have long served as standard but computationally heavy wrappers for instance segmentation during inference. That is, we propose an end-to-end differentiable solution to the nuclei segmentation problem. Given sufficiently powerful hardware, the method can, in principle, process a complete WSI at once. However, due to our current hardware limitations, we still divide the WSI into large tiles for inference.

\paragraph{Compact Polygon Representation and Flexible-Range Radial Loss}
Representing every possible nucleus by a separate dense pixel mask scales with both image area and the number of instances, making it costly for dense, high-resolution WSIs. A compact shape descriptor allows instances to occupy the same image region without imposing a hard per-pixel assignment. Among alternatives such as Fourier descriptors and splines, star-convex polygons are particularly well-suited to our objective because every boundary direction is represented by an explicit radial distance. This makes it possible to supervise each ray by a geometrically derived interval; global, coupled shape descriptors do not admit the same direct ray-wise range loss. Furthermore, StarDist has demonstrated that star-convex polygons accurately and efficiently represent nuclear morphology~\cite{StarDist,StarDistClassification}. Building on this representation, we introduce a novel radial distance loss that supervises each predicted distance against a plausible range rather than a single target. This is critical because common datasets lack pixel-level occlusion annotations, meaning that traditional loss functions force models to learn rigid, artificial, non-overlapping boundaries, leading to unrealistic nuclei shapes wherever cells physically touch or overlap. Our range acts as an inductive bias toward biologically plausible geometry, allowing realistic overlaps between nuclei to emerge naturally from standard, non-overlapping masks. To the best of our knowledge, LSP-DETR is the first method to model biologically realistic nuclei overlaps without requiring explicit overlap annotations during training. We detail the formulation in \cref{sec:loss}.

\paragraph{Algorithmic Efficiency and Input-Size Agnosticism}
LSP-DETR achieves linear computational complexity and input-size agnosticism, allowing it to process exceptionally large tiles in a single forward pass despite being trained on standard $256\times256$ px patches. These properties follow from two design choices, relative rotary positional encodings and local neighborhood attention, detailed in \cref{sec:method:layer}. Although hardware limitations still prevent processing an entire whole-slide image at once, the method delivers an inference speed of $0.45 \text{ s/mm}^2$ on the TCGA cohort, a $3.2\times$ speedup over the next-fastest state-of-the-art model, StarDist.

\paragraph{Overlap-Aware Evaluation}
The annotation mismatch is not only a training problem but also an evaluation problem: standard Panoptic Quality (PQ) penalizes an overlap-capable model for predicting plausible regions assigned exclusively to a neighboring instance in the ground truth. We therefore introduce Masked Panoptic Quality (MPQ), which evaluates a matched prediction without penalizing its overlap with other annotated nuclei. We complement this metric with a targeted two-pathologist protocol: the pathologists classified whether each neighboring-nucleus pair was merely touching or genuinely overlapping. An initial review covered 600 pairs, and a second pathologist independently assessed a stratified subset of 210 pairs. On the 115 pairs for which both pathologists reached a diagnostic consensus, the predicted polygon-intersection area distinguished touching from overlapping nuclei with an IPW-weighted F$_1$-score of 0.964 and an AUROC of 0.983, supporting the alignment of the learned overlaps with expert assessment.

\paragraph{Cross-Dataset Generalization}
Finally, we evaluate robustness under direct transfer from PanNuke to MoNuSeg without retraining or adaptation. LSP-DETR reaches an F$_1$-score of 85.0 on MoNuSeg, outperforming the substantially larger LKCell model and demonstrating robustness across datasets and image resolutions.

\subsection{Related Work}

Various methods have been developed for nuclei instance segmentation, ranging from classical image-processing approaches to advanced automated deep-learning techniques.
Despite these advances, standard deep-learning models traditionally rely on indirect strategies for instance segmentation, such as two-stage detectors or one-stage detectors with post-processing, as discussed below. While specialized DETR-like models offer direct set prediction, their reliance on absolute positional encodings and quadratic attention often restricts them to fixed image sizes, preventing a truly flexible, canonical solution.

\subsubsection{Two-Stage Detectors}

Two-stage methods typically begin by generating region proposals via bounding box prediction, followed by a segmentation stage that produces the final masks.
Mask~R-CNN~\cite{he2017mask} is the most prominent example of this category.
Vuola et al.~\cite{vuola2019mask} combined Mask~R-CNN with U-Net~\cite{unet}, leveraging the strengths of both architectures to improve nuclei segmentation accuracy.
Similarly,~\citet{bancher2021improving} augmented Mask~R-CNN by including distance maps predicted by U-Net as additional inputs, enabling more precise nuclei separation.

Other works have extended the two-stage framework with additional stages to refine segmentation.
For instance,~\citet{7562400} proposed a three-stage approach consisting of segmentation, cell labeling, and cell boundary refinement.
Although effective, this method is very slow and is not designed for end-to-end training, making it less practical for large-scale applications.

While effective, these methods are computationally complex and often require additional post-processing steps to handle overlapping nuclei and deduplication.

\subsubsection{One-Stage Detectors with Post-Processing}

One-stage detectors simplify the segmentation process by using a single network for prediction, with post-processing techniques to resolve overlapping nuclei and ensure instance separation.
DIST~\cite{8438559}, for instance, employs a U-Net-style fully convolutional network (FCN) to predict a distance map in which each pixel's value represents its distance to the nearest background pixel.
The resulting distance map is post-processed using a watershed algorithm.
HoVer-Net~\cite{graham2019hover} introduced a multi-branch decoder network that predicts horizontal and vertical distance maps alongside a binary mask and class mask.
These outputs are processed using the watershed algorithm to create nuclei instances.

Building on this approach, CellViT~\cite{CellViT} improved Ho\-Ver-Net by incorporating the Segment Anything Model (SAM)~\cite{kirillov2023segany} for enhanced feature extraction. Further advances were made by LKCell~\cite{LKCell}, which unified the decoder branches and replaced the computationally expensive SAM with large convolutional kernels~\cite{ding2024unireplknet}, achieving superior accuracy. Most recently, $\text{CellViT}^{++}$~\cite{horst2025cellvit++} explored replacing the SAM feature extractor with pathology foundation models. While this approach demonstrated successful data-efficient adaptation for segmentation and classification, the original CellViT using SAM-H still maintained the highest performance across most benchmarks.

Another category of one-stage detectors focuses on predicting shape descriptors for each nucleus, describing them as polygons.
StarDist~\cite{StarDist}, a popular approach in this domain, predicts radial distances at equidistant angles from each pixel to the nucleus boundary to form a star-convex polygon.
It uses a Euclidean distance map to the nearest background pixel, with non-maximum suppression (NMS) to deduplicate overlapping predictions.
This method has also been extended for nuclei classification~\cite{StarDistClassification}.
CPP-Net~\cite{chen2023cpp} improves StarDist by refining the prediction of radial distances using a weighted average of radial distance predictions from a pixel's neighbors along the distance rays, leading to more accurate results.
SplineDist~\cite{SplineDist}, in contrast, replaces radial distances with splines to describe nuclei contours.
While splines offer advantages like enforced smoothness that better match the natural shapes of nuclei, they lack an orthogonal basis, complicating the loss calculation as differently parametrized splines can represent the same shape.
To address this limitation, FourierDist~\cite{FourierDist} employs Fourier descriptors, which have an orthogonal basis and can accurately capture nuclei shapes with fewer parameters than star-convex polygons.

Although these methods are highly effective and achieve state-of-the-art performance on most nuclei segmentation benchmarks, the reliance on per-pixel prediction and handcrafted post-processing methods slows down inference, especially when working with large input images.
Additionally, StarDist-based methods struggle with predicting overlapping nuclei due to NMS, while HoVer-Net-based methods cannot predict overlapping nuclei at all.

\subsubsection{DETR-Based Approaches}
\label{sec:detr}

The closest approaches to set prediction are autoregressive sequence models, such as recurrent neural networks and transformers.
While these models can handle sequential data, they present batching challenges and inference times that increase in proportion to the sequence length.

To overcome these limitations, the DEtection TRansformer (DETR)~\cite{detr} introduces a parallel set prediction approach. DETR employs a fixed set of learnable object queries whose quantity exceeds the maximum number of objects typically found in an image. The network predicts objects for all queries simultaneously but learns to filter the surplus by classifying them as \say{no object}. This mechanism allows for direct, end-to-end prediction, effectively eliminating the need for heuristic post-processing steps.

The adoption of DETR-based methods in nuclei instance segmentation is an emerging research direction.
Cell-DETR~\cite{prangemeier2020c} was the first to adapt DETR for this domain, demonstrating yeast-cell segmentation in microstructured microscopy by predicting a binary mask for each query to represent individual cells.
NucDETR~\cite{obeid2022nucdetr} explored a similar approach but limited its predictions to bounding boxes.
ACFormer~\cite{Huang_2023_ICCV} focuses on predicting nuclei centroids using an affine-consistent transformer framework.
This method incorporates a local-global network architecture, where the local network predictions are guided by the global network's broader contextual understanding, ensuring scale consistency across varying image magnifications.

The most relevant approach to ours is SAP-DETR~\cite{liu2023sap}, developed for general object detection using bounding boxes.
It introduces query-specific positions by distributing object queries over a fixed mesh grid while allowing dynamic movement of queries.
Guided by specific characteristics of nuclei, our method extends this concept by eliminating the global-context requirement, thereby enabling dynamic scaling of object queries based on input size and resolution.

\section{Method}
\subsection{LSP-DETR Model}

At a high level, the model receives an image as input and outputs a set of shape descriptors for segmented nuclei, together with their semantic embeddings and classes.
Each nucleus is represented by its \emph{shape descriptor}, which consists of a position vector $p\in[0,1]^{2}$ (relative image coordinates) and a sequence $r\in\mathbb{R}_{+}^{64}$ representing radial distances from $p$ along 64 evenly distributed rays.
These polygons can approximate any star-convex region.

Initially, we cover the input image with descriptors approximating circles and distribute them evenly in a regular grid.
The algorithm then iteratively adjusts the descriptors so that each segments a nucleus within its neighborhood.

The architecture consists of a feature extractor (a backbone in DETR terminology) that processes the image and a transformer module that operates on shape descriptors and embeddings.
The whole architecture is outlined in Figure~\ref{fig:model:overview}.

\subsubsection{Visual Feature Extractor}

Given a square input image of dimensions $3 \times R \times R$, the backbone extracts several multi-scale features $\mathcal{F}_{i}$ of size $C_{i}\times R_{i}\times R_{i}$ and a final feature map $B$, with dimensions $C_{B}\times R_{B}\times R_{B}$.
The entire architecture can be readily generalized to rectangular images, but we restrict ourselves to the square case for simplicity.

We adopt a transformer-based backbone, motivated by the observation that conventional CNNs, which rely on padding, can introduce minor positional inaccuracies in radial-distance estimates when applied to images of varying resolutions.
In particular, we employ the widely used and efficient Swin-V2-T~\cite{liu2021swinv2}, a lightweight transformer backbone that is agnostic to input image size.
Features $\mathcal{F}_{1}$, $\mathcal{F}_{2}$, $\mathcal{F}_{3}$, and output~$B$ correspond to embedding vectors taken after Swin stages 1--4.
We fine-tune the model starting from weights pretrained on ImageNet-1k.
However, the model is in principle compatible with any locally based backbone architecture.

\begin{figure}[!t]
    \centering
    \input{fig/model/overview.tex}
    \caption{A Swin-V2-T backbone extracts multi-scale features $\mathcal{F}_{1\text{--}3}$ and a final feature map $B$. Object queries are initialized on a regular grid of circular shape descriptors $(p^0,r^0)$, with embeddings $e^0$ obtained by sampling $B$ at the corresponding positions. The queries are iteratively refined through $L$ transformer layers to produce class, position, and radial-shape predictions; light-blue descriptors denote background queries.}
    \label{fig:model:overview}
\end{figure}

\begin{figure}[!t]
    \centering
    \begin{tikzpicture}[
    node distance=0.2cm and 0.2cm,
    every node/.style={text=our-ink, font=\small, inner ysep=2pt},
    base_block/.style={rectangle, rounded corners=2pt, draw=our-ink, align=center},
    norm_block/.style={base_block, fill=our-baby, text width=8.1em, inner ysep=2pt},
    w_block/.style={base_block, fill=our-pink, inner xsep=2pt, text depth=0.4ex},
    label/.style={text=our-light-ink, font=\scriptsize},
    >={Triangle[length=3pt, width=4pt]},
    sta_block/.style={base_block, fill=our-sta},
    ffn_block/.style={base_block, fill=our-yellow, text width=8.1em, inner ysep=2pt},
    pe_block/.style={base_block, fill=our-green},
    mlp_block/.style={base_block, fill=our-light-yellow, text width=2em, inner ysep=2pt},
    container/.style={rectangle, rounded corners=10pt, draw=black, fill=our-bg, dashed, dash pattern=on 2pt off 2pt},
    attn_container/.style={container, rounded corners=5pt, fill=our-layer},
    over_attn_bg/.style={preaction={draw, our-layer, line width=#1}, double=none},
    over_attn_bg/.default=3pt,
    over_main_bg/.style={preaction={draw, our-bg, line width=#1}, double=none},
    over_main_bg/.default=3pt
]

\node (e_input) {$e^{\ell-1}$};
\node[right=1.4cm of e_input] (p_input) {$p^{\ell-1}$};
\node[right=-0.2cm of p_input] (r_input) {$r^{\ell-1}$};


\node[w_block, above=1.1cm of e_input] (s_wk) {$W_K$};
\node[w_block, right=of s_wk] (s_wq) {$W_Q$};
\node[w_block, left=of s_wk] (s_wv) {$W_V$};

\path let \p1=(s_wv.west), \p2=(s_wq.east) in node[sta_block, minimum width=\x2-\x1, above=1.3cm of s_wk] (s_sta) {STA};

\draw (s_wq.south) -- ++(0, -0.15cm) -| (s_wv.south);
\draw (e_input) -- (s_wk.south -| e_input.north);
\draw[->] (s_wq.north) -- (s_sta.south -| s_wq.north) node[pos=0.8, right] {$Q$};
\draw[->] (s_wk.north) -- (s_sta.south -| s_wk.north) node[pos=0.8, right] {$K$};
\draw[->] (s_wv.north) -- (s_sta.south -| s_wv.north) node[pos=0.8, right] {$V$};

\path let \p1=(s_wk.west), \p2=(s_wq.east) in node[pe_block, minimum width=\x2-\x1, anchor=west, at={([yshift=0.5cm]s_wk.north west)}] (s_pe) {PE};

\node[above=0.6cm of s_sta] (add_node1) {$\oplus$};

\draw[->, shorten >=-3pt] ($(s_wk.south) - (0, 0.6cm)$) -- ++(-2cm, 0) |- (add_node1.west);

\node[norm_block, above=0.1cm of add_node1] (norm1) {Norm};
\draw[->] (s_sta.north) -- (norm1.south);

\draw[->] (p_input |- s_pe) |- (s_pe.east);


\node[w_block, above=0.9cm of norm1] (c_wk) {$W_K$};
\node[w_block, right=of c_wk] (c_wq) {$W_Q$};
\node[w_block, left=of c_wk] (c_wv) {$W_V$};

\path let \p1=(c_wv.west), \p2=(c_wq.east) in node[sta_block, minimum width=\x2-\x1, above=1.3cm of c_wk] (c_sta) {STA};

\draw[->] (c_wq.north) -- (c_sta.south -| c_wq.north) node[pos=0.8, right] {$Q$};
\draw[->] (c_wk.north) -- (c_sta.south -| c_wk.north) node[pos=0.8, right] {$K$};

\node[pe_block, text width=1.4em, above=0.5cm of c_wq.center] (c_pe_q) {PE};
\node[pe_block, text width=1.4em, above=0.5cm of c_wk.center] (c_pe_k) {PE};

\node[above=0.6cm of c_sta] (add_node2) {$\oplus$};
\node[norm_block, above=0.1cm of add_node2] (norm2) {Norm};

\draw[->] (c_sta.north) -- (norm2.south);

\draw[->] (p_input |- c_pe_q) |- (c_pe_q.east);

\node[left=1.55cm of c_wv.south, yshift=-0.15cm] (feature) {$\mathcal{F}^{\ell}$};
\node (feature_pos) at ([xshift=0.09cm]feature |- c_pe_k) {$t^{\ell}$};
\draw (feature) -| (c_wv.south) (feature) -| (c_wk.south);
\draw[->] (feature_pos) -- (c_pe_k);

\draw[->, over_main_bg, shorten >=-3pt] ($(norm1) + (0, 0.5cm)$) -- ++(-2cm, 0) |- (add_node2.west);
\draw (norm1) -- ++(0, 0.5cm) -| (c_wq);
\draw[->, over_attn_bg] (c_wv.north) -- (c_sta.south -| c_wv.north) node[pos=0.8, right] {$V$};


\node[ffn_block, above=0.6cm of norm2] (ffn) {FFN};
\draw[->] (norm2) -- (ffn);

\node[above=0.1 of ffn] (add_node3) {$\oplus$};

\node[norm_block, above=0.1cm of add_node3] (norm3) {Norm};
\draw[->] (ffn) -- (norm3);
\draw[->, shorten >=-3pt] ($(norm2) + (0, 0.5cm)$) -- ++(-2cm, 0) |- (add_node3.west);

\begin{scope}[on background layer]
    \path let
        \p1 = (norm3.west),
        \p2 = (norm3.north),
        \p3 = (s_wq.south),
        \p4 = (r_input.east)
    in
    \pgfextra{
      \coordinate (box-nw) at (\x1-0.63cm, \y2+1.3cm); 
      \coordinate (box-se) at (\x4-0.0cm, \y3-0.9cm); 
    };
    \node[container, fit=(box-nw) (box-se), inner sep=0] (box) {};

    \path let
        \p1 = (s_sta.west),
        \p2 = (s_sta.north),
        \p3 = (s_wq.south),
        \p4 = (s_sta.east)
    in
    \pgfextra{
      \coordinate (self-attn-box-nw) at (\x1-0.3cm, \y2+0.5cm); 
      \coordinate (self-attn-box-se) at (\x4+0.3cm, \y3-0.3cm); 
    };
    \node[attn_container, fit=(self-attn-box-nw) (self-attn-box-se), inner sep=0] (self_attn_box) {};

    \path let
        \p1 = (c_sta.west),
        \p2 = (c_sta.north),
        \p3 = (c_wq.south),
        \p4 = (c_sta.east)
    in
    \pgfextra{
      \coordinate (cross-attn-box-nw) at (\x1-0.3cm, \y2+0.5cm); 
      \coordinate (cross-attn-box-se) at (\x4+0.3cm, \y3-0.3cm); 
    };
    \node[attn_container, fit=(cross-attn-box-nw) (cross-attn-box-se), inner sep=0] (cross_attn_box) {};
\end{scope}
\node[label, anchor=north west, inner sep=4pt] at (self_attn_box.north west) {Self-Attn};
\node[label, anchor=north west, inner sep=4pt] at (cross_attn_box.north west) {Cross-Attn};

\node (e_output) at ($(e_input |- box.north) + (0, 0.5cm)$) {$e^\ell$};
\node (p_output) at (p_input |- e_output) {$p^\ell$};
\node (r_output) at (r_input |- e_output) {$r^\ell$};
\node (c_output) at ([xshift=-0.8cm]e_output) {$c^\ell$};

\draw[->] (norm3) -- (e_output.south);
\draw[->] (p_input) -- (p_output.south);
\draw[->] (r_input) -- (r_output.south);
\draw[->] ([yshift=0.65cm]norm3.north) -| (c_output.south);
\node[w_block] at ([yshift=0.65cm]norm3.north -| c_output) {$W_c$};

\node (add_node_r) at ([yshift=0.6cm] norm3 -| r_output) {$\oplus$};
\node (add_node_p) at ([yshift=1.1cm] norm3 -| p_output) {$\oplus$};

\draw[->, shorten >=-3pt] (norm3) |- (add_node_r.west);
\draw[->, shorten >=-3pt] (norm3) |- (add_node_p.west);

\node[mlp_block] (mlp_p) at ([xshift=0.9cm] norm3 |- add_node_p) {MLP};
\node[mlp_block] (mlp_r) at ([xshift=0.9cm] norm3 |- add_node_r) {MLP};

\draw[over_main_bg] (norm3 -| add_node_p) -| (add_node_p.south);

\node[
    label,
    anchor=north west,
    inner sep=6pt
] at (box.north west) {Layer $\ell$};

\end{tikzpicture}
    \caption{Each transformer layer applies local STA-based self-attention between object queries, cross-attention with image features $\mathcal{F}^{\ell}$, and an FFN. Rotary positional encodings incorporate query and feature coordinates into the attention mechanism, while separate heads predict classes $c^\ell$, positions $p^\ell$, and radial distances $r^\ell$.}
    \label{fig:model:layer}
\end{figure}

\subsubsection{Transformer Module}

The transformer operates on $N$ \emph{object queries} $(e^{\ell}_{j}, p^{\ell}_{j} , r^{\ell}_{j})$ for $j=1,\ldots, N$, with the superscript $\ell$ indexing the layer.
For a query $j$, the vector $e^{\ell}_{j}\in\mathbb{R}^{d}$ denotes a content embedding, $p^{\ell}_{j}\in[0,1]^{2}$ is a relative position in the image, and $r^{\ell}_{j}\in\mathbb{R}_{+}^{64}$ are radial distances approximating the nucleus shape.
We selected 64 rays as in StarDist~\cite{StarDistClassification}.
We also collect all query embeddings into $e^{\ell}=(e^{\ell}_{1},\ldots,e^{\ell}_{N})$ and proceed similarly for~$p^{\ell}$ and~$r^{\ell}$.

The initial nuclei descriptors $(p^{0}, r^{0})$ are set to approximations of circles of radius~$s$ distributed evenly in a regular grid and tightly packed in the image, as depicted in Figure~\ref{fig:model:overview}.
The hyperparameter~$s$ depends on the image size~$R$ (px), resolution (\textmu m/px), and the biological characteristics of nuclei, roughly representing the relative minimum nucleus radius.
While the minimum thickness of highly flattened nuclei can be as small as approximately 3~\textmu m~\cite{lammerding2011mechanics}, we set the initial grid cell diameter to 3.5~\textmu m to ensure adequate separation and localization.
The normalized radius $s$ is thus defined as
\begin{equation} \label{eq:s}
    s = \frac{3.5\,\text{\textmu m}}{2 \times R\times\text{resolution}}.
\end{equation}

To define~$e^{0}$, we take the backbone output~$B$ and evaluate it at the positions~$p^{0}$.
Specifically, let $B(p^{0}_j)\in\mathbb{R}^C$ be the pixel value of the backbone output $B$ at the relative position $p^{0}_j$, obtained using bilinear interpolation between neighboring pixels.
We then define $e^{0}=\operatorname{Norm}\bigl(W B(p^{0})\bigr)$, where $W \in \mathbb{R}^{d \times C}$ is a linear transformation and $\operatorname{Norm}$ is a layer normalization.

The object queries are iteratively refined by the \emph{transformer layers} in the \emph{transformer decoder}; see Figure~\ref{fig:model:overview}.
During this process, the model adjusts the nuclei positions and shapes and updates the information collected in the embeddings.
The architecture is designed so that each object query can segment the nucleus in its vicinity.
Therefore, the positions $p^\ell$ are restricted to remain in the grid cell in which they were initiated.

Moreover, as the number of nuclei is overapproximated, the model must have the ability to disregard the entire query.
Therefore, we employ a linear classifier that predicts nuclei classes $c^{\ell}$ from their embeddings $e^{\ell}$ after each layer $\ell$ of the transformer.
The number of classes depends on the task, but there is always at least one class reserved for the \say{no nucleus} category, denoted by $\varnothing$, to drop the query in the final set of segmented nuclei.

\subsubsection{Transformer Layers}
\label{sec:method:layer}
Our decoder architecture makes two fundamental modifications to the standard transformer layers to improve both efficiency and geometric awareness.
First, we replace absolute positional encodings with a unified relative encoding scheme, using the multidimensional Cayley-STRING rotary encodings~\cite{schenck2025learning}.
This integrates relative spatial information into the attention dot product by rotating the query and key vectors.
Second, recognizing that segmentation is an inherently local operation, we replace the standard quadratic-cost full attention with Sliding Tile Attention (STA)~\cite{zhang2025fast}.
This mechanism efficiently restricts attention to local neighborhoods, achieving linear complexity.
The layer design is outlined in Figure~\ref{fig:model:layer}.

\paragraph{Self-Attention}
In these layers, each object query is refined by attending to the object queries in its local neighborhood.
This neighborhood remains constant across all transformer layers because the positions are updated to stay within the grid cell in which they were initialized.
Consequently, the neighborhood is determined by the initial positions $p^0$ of the object queries.
STA is used to efficiently implement this local attention mechanism, providing sufficient context for queries to interact and perform tasks such as deduplication.

To encode relative positions (PE), we use Rotary Positional Encodings (RoPE).
The query $Q_j$ and key $K_j$ vectors are obtained from the embedding $e^{\ell-1}_j$ of the object query by applying a rotational matrix corresponding to the position $p_j^{\ell-1}$ as
\begin{equation} \label{eq:rope_query}
    Q_j = \operatorname{RoPE}(p^{\ell-1}_j / s) P^\ell W_Q e^{\ell-1}_j,
\end{equation}
and
\begin{equation}
    K_j = \operatorname{RoPE}(p^{\ell-1}_j / s) P^\ell W_K e^{\ell-1}_j.
\end{equation}
Here, $P^\ell$ is a learnable orthogonal transformation shared by $Q_j$ and $K_j$ but unique for each layer, and RoPE is the two-dimensional rotary positional encoding used in Cayley-STRING~\cite{schenck2025learning}.
Note that the positions are divided by~$s$ to make the encoding scale-independent.

\paragraph{Cross-Attention}
In cross-attention layers, object queries interact with image features. This interaction is also restricted to a fixed neighborhood defined by the queries' initial positions. The query vector $Q_j$ is derived from the intermediate embeddings produced by self-attention, with positional encoding based on $p^{\ell-1}$.

In contrast, the keys $K_j$ and values $V_j$ are derived from the image feature~$\mathcal{F}^\ell$.
The multi-scale image features are fed from the backbone into the decoder in a round-robin fashion, progressing from the lowest to the highest resolution; that is, we set $\mathcal{F}^\ell=\mathcal{F}_{3- ( (\ell-1) \bmod 3)}$ in our setting.
However, we must convert the image features to the same format as the object queries.
Therefore, we transform the image feature $\mathcal{F}^\ell$ of size $C^\ell\times R^\ell\times R^\ell$ to a sequence of embeddings $f^\ell_j\in\mathbb{R}^{C^\ell}$ and relative image positions $t^\ell_j\in[0,1]^2$.

To this end, we fix a one-to-one mapping (enumeration) between indices $j$ ($0\le j< (R^\ell)^2$) and feature positions $(x,y)$ for $0\le x,y<R^\ell$.
Namely, we take $j=R^\ell y + x$, corresponding to flattening.
Then we set the key token as the local image feature $f^\ell_j = \mathcal{F}^\ell(x, y)\in\mathbb{R}^{C^\ell}$
and its relative image position as $t^\ell_j = \tfrac{1}{R^\ell}(x+\tfrac12, y+\tfrac12)$.
Finally, the keys $K_j$ are given by
\begin{equation}
    K_j = \operatorname{RoPE}(t^\ell_j / s) P^\ell W_K f^\ell_j,
\end{equation}
where, as in self-attention, the learnable orthogonal transform~$P^\ell$ is shared by queries and keys but is unique to each layer.
This encoding strategy ensures that object queries and image features share a consistent spatial representation, which is critical for aligning them accurately during the attention process.

\paragraph{Prediction Heads}
At each layer $\ell$, the output embedding $e^{\ell}$ of each object query is passed to two prediction heads to produce the position~$p^{\ell}$ and radial distances~$r^{\ell}$.
A three-layer MLP predicts an offset $\Delta p^{\ell}\in \mathbb{R}^{2}$.
The positions~$p^\ell$ are then updated from~$p^{\ell-1}$ similarly to SAP-DETR~\citep{liu2023sap}; that is,
\begin{equation} \label{eq:position-update}
    p^{\ell}=s\tanh\bigl(\tanh^{-1}\bigl((p^{\ell-1}-p^{0})/s\bigr) + \Delta p^{\ell} \bigr) + p^{0},
\end{equation}
ensuring that the nucleus does not drift by more than $s$ along any coordinate axis from its original position $p^0$.
Another three-layer MLP predicts a log-scale adjustment $\Delta r^\ell\in\mathbb{R}^{64}$, and the radial distances are updated multiplicatively as $r^\ell=r^{\ell-1}\exp(\Delta r^\ell)$.

\subsubsection{Set Prediction Loss}
\label{sec:loss}

Our model formulates instance segmentation as a direct set prediction task, outputting final instances without the need for post-processing.
Consequently, the loss computation is a two-step process: first, an optimal bipartite matching procedure assigns each ground-truth nucleus to the best-fitting prediction; second, a loss function penalizes the dissimilarities within these assigned pairs. For this loss, we employ the \say{look forward twice} scheme from~\cite{zhang2022dino}. Because the loss and matching processes are performed at each decoder layer, we omit the layer index $\ell$ from the predictions in this section for brevity.

The total loss is designed to optimize two primary objectives: accurate nuclei classification and high-fidelity shape reconstruction.
We encourage faithful shape modeling by training the network to predict the points $p$ near the nucleus centroid.
This approach is based on the assumption that radial distances from a central point are more likely to parameterize nuclei shapes effectively.

To match predictions to ground-truth instances, we adopt the bipartite matching strategy used in DETR.
Let the ground-truth set consist of $M$ nuclei, where the $j$-th nucleus is represented by its class label~$\overline{c}_j$ and binary instance mask $\overline{m}_j \in \{ 0, 1 \}^{R \times R}$.
From the mask, we compute the ground-truth nucleus centroid $\overline{p}_j$.
The model outputs $N$ predictions, where $N$ corresponds to the number of grid cells and therefore varies with the input image's dimensions.
The grid density is chosen to ensure $N \geq M$. Each prediction $j$ consists of a class score $c_j$, predicted point $p_j$, and a vector of radial distances~$r_j$.

\paragraph{Bipartite Matching}
The core of our method is to find an optimal one-to-one assignment between the $M$ ground-truth nuclei and the $N$ predictions.
We find the optimal assignment $\sigma^*$ over all $M$-permutations of $N$ elements $\sigma \colon \{ 1, \ldots, M \} \to \{1, \ldots, N \}$ by minimizing the pairwise matching cost
\begin{align}
	\mathcal{W}(\sigma)
		& = \sum_{j=1}^{M}
			\mathcal{W}_c (\overline{c}_{j}, c_{\sigma(j)})
			+ \mathcal{L}_{p}(\overline p_{j}, p_{\sigma(j)})
            \\
			&\qquad+ \mathcal{L}_r (p_{\sigma(j)}, r_{\sigma(j)})
				+ \mathcal{W}_m (\overline m_j, p_{\sigma(j)}),
\end{align}
where the classification loss $\mathcal{W}_c$ is the focal-loss adopted exactly from~\cite{zhang2022dino}, including its hyperparameters.
The term $\mathcal{L}_p$ is the~$L^1$ distance between the ground-truth nucleus centroid~$\bar p_j$ and the predicted point~$p_{\sigma(j)}$.
Next, $\mathcal L_r$ is our proposed radial distance loss, detailed below, and $\mathcal W_m$ is an inner mask cost that discourages predictions of positions outside the nucleus, defined by
\begin{equation}
  \mathcal{W}_m (m,p)
		= \begin{cases}
				0
					& \text{if the point $p$ lies inside the mask $m$},
					\\
				\lambda
					& \text{otherwise}.
			\end{cases}
\end{equation}
The weight $\lambda$ is set so that the misassignment of the position $p_{\sigma(j)}$ to the ground-truth mask $\overline m_j$ dominates the other cost components; for example, $\lambda=10$.
The optimal assignment is found efficiently using a GPU-accelerated linear assignment solver~\cite{anonymous2024hotpp}.

\paragraph{Loss Function}
After determining the optimal assignment $\sigma^*$, the final loss sums shape-reconstruction losses over the matched pairs and a classification loss over all predictions:
\begin{align}
    \begin{split}
      \mathcal{L}
        & = \frac{1}{N}\sum_{j=1}^{M} \mathcal{L}_c(\overline c_j, c_{\sigma^*(j)})
            + \frac{1}{N}\sum_{\substack{1\le k\le N \\ k\notin\operatorname{Range}(\sigma^*)}}
                \mathcal{L}_c(\varnothing, c_{k})
            \\
        &\quad + \frac{1}{M}\sum_{j=1}^{M}
            \left[
                \mathcal{L}_p(\overline p_j, p_{\sigma^*(j)})
              + \mathcal{L}_r(p_{\sigma^*(j)}, r_{\sigma^*(j)})
            \right].
  \end{split}
\end{align}
The classification loss $\mathcal L_c$ is the standard focal loss~\cite{focal}.
It is applied to all $N$ predictions so that for the $M$ matched predictions, the target is the corresponding ground-truth class $\overline c_j$, and for the remaining $N-M$ unmatched predictions ($k\notin\operatorname{Range}(\sigma^*)$), the target is the \say{no nucleus} ($\varnothing$) class.
The point loss $\mathcal L_p$ and radial loss $\mathcal L_r$ are computed only for the $M$ matched pairs.

\paragraph{Radial Distance Loss}
The radial distance loss $\mathcal L_r$ is designed to supervise radial-distance prediction effectively while addressing several key challenges.
A naive approach, such as a direct $L^1$ loss between predicted distances and a single ground-truth vector, is impractical.
First, computing precise ground-truth distances on the fly or storing them in advance is computationally expensive and memory-intensive.
Second, in domains such as nuclei segmentation, ground-truth masks often contain overlapping instances for which the overlapping regions are not explicitly annotated.
A standard loss would incorrectly penalize a model for predicting these physically plausible overlaps.
Our approach avoids this by design.

To overcome these limitations, we propose a loss that penalizes predictions only when they fall outside a plausible range, rather than enforcing a single exact value.
This range is defined by a lower and an upper bound.
Crucially, these bounds and the subsequent loss are computed only for positions~$p_j$ that lie inside any ground-truth nucleus mask~$\overline{m}$.
We call these positions \emph{foreground points}.
All remaining points are ignored by the loss function.

\begin{figure*}[tb]
    \centering
    \begin{subfigure}[b]{0.22\textwidth}
        \centering
        \includegraphics[width=\linewidth]{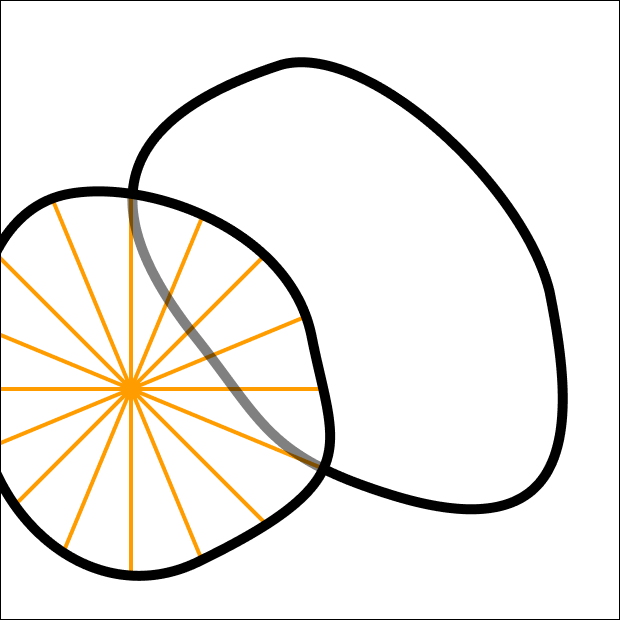}
        \caption*{$r_{\min}$}
    \end{subfigure}
    \hspace{0.5em}
    \begin{subfigure}[b]{0.22\textwidth}
        \centering
        \includegraphics[width=\linewidth]{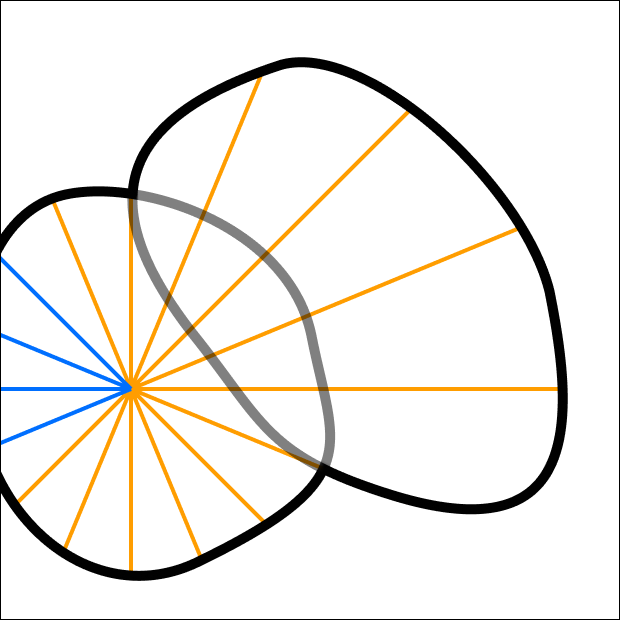}
        \caption*{$r_{\max}$}
    \end{subfigure}
    \hfill
    \begin{subfigure}[b]{0.22\textwidth}
        \centering
        \includegraphics[width=\linewidth]{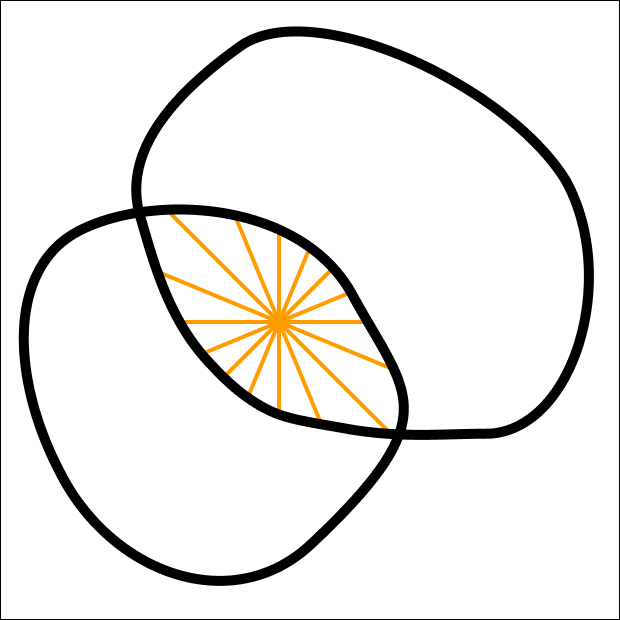}
        \caption*{$r_{\min}$}
    \end{subfigure}
    \hspace{0.5em}
    \begin{subfigure}[b]{0.22\textwidth}
        \centering
        \includegraphics[width=\linewidth]{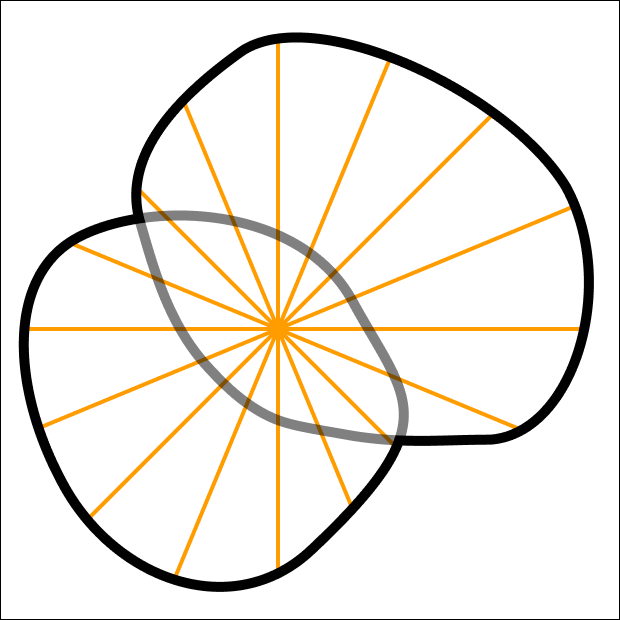}
        \caption*{$r_{\max}$}
    \end{subfigure}
    \caption{%
        Visualization of the lower bound $r_{\min}$ and upper bound $r_{\max}$ for radial distances originating from a point $p$. 
        Ground-truth nuclei are delineated by black outlines. 
        The scenarios are organized into two groups: (left) a point within a nucleus near the image boundary, and (right) a point within an overlapping region.
        Blue rays indicate an infinite distance bound, which occurs when a ray does not intersect a background pixel within the image frame.
    }
    \label{fig:radial-distances}
\end{figure*}

Let $r$ be the vector of 64 predicted radial distances from a foreground point~$p$.
For this point, we define the \emph{lower bound}~$r_\text{min}\in\mathbb{R}^{64}$ as a vector of radial distances from~$p$ to the nearest edge of any ground-truth mask containing~$p$, or to the image boundary.
Next, we set the \emph{upper bound}~$r_\text{max}\in\mathbb{R}^{64}$ as a vector of radial distances from~$p$ to the nearest pixels that are not covered by any ground-truth nucleus mask.
If the whole ray is covered by the ground-truth masks up to the image boundary, we set~$r_\text{max}$ to infinity.
For illustration, see~\cref{fig:radial-distances}.

Finally, the loss is zero for any predicted distance that lies within its respective bounds $[r_\text{min}, r_\text{max}]$ and incurs a penalty proportional to the violation otherwise.
Namely, we set
\begin{equation}
    \mathcal L_r (p, r)
        = \frac{1}{64} \sum_{k=1}^{64}
            \max\left\{
                r_\text{min} - r,
                0,
                r - r_\text{max}
            \right\}_k,
\end{equation}
where the maximum is computed elementwise.
This design is highly efficient because both the lower and upper bound vectors can be precomputed for each pixel, independent of the number of instances in an image. 
Furthermore, it robustly handles object overlaps and annotation ambiguities by only penalizing demonstrably incorrect predictions.

\subsubsection{Evaluation Metrics}

We assess model performance on two distinct tasks: instance segmentation and instance detection.
For consistency and a fair comparison, we adopt the primary metrics and matching criteria used in CellViT~\cite{CellViT}.

Let $\overline m_1, \dots, \overline m_M$ denote the ground-truth masks. Similarly, let $m_1, \dots, m_N$ represent the instance masks obtained by rasterizing predicted star-convex polygons, and let $c_1, \ldots, c_N$ denote their predicted classes from $K$ classes.

\paragraph{Instance Segmentation}

We quantify instance segmentation quality using a unified metric called \emph{Panoptic Quality}~(PQ)~\cite{kirillov2019panoptic}, which captures both segmentation and detection quality.
The evaluation process begins by matching predictions to ground-truth instances of the same class.
For each class, we find an optimal bipartite matching between the sets of ground-truth and predicted masks that maximizes the total intersection over union (IoU).

Based on the predicted and ground-truth instances, we define a matched pair $(\overline m, m)$ as one where $\operatorname{IoU}(\overline m,m)>0.5$.
We define \emph{True Positives}~(TP) as the set of all such matched pairs, \emph{False Positives}~(FP) as the set of predicted instances $m$ that are not part of any matched pair, and \emph{False Negatives}~(FN) as the ground-truth instances $\overline m$ that are not part of any matched pair.
For a class $c$, we compute \emph{Segmentation Quality}~(SQ$_c$) as 
\begin{equation}
    \text{SQ}_c
        = \frac{1}{|\text{TP}|}
            \sum_{(\overline m, m) \in \text{TP}}
                \operatorname{IoU}(\overline m, m),
\end{equation}
and \emph{Recognition Quality}~(RQ$_c$) as
\begin{equation}
    \text{RQ}_c
        = \frac
            {|\text{TP}|}
            {|\text{TP}|
                + \frac{1}{2}|\text{FP}|
                + \frac{1}{2}|\text{FN}|}.
\end{equation}
Finally, Panoptic Quality is their product $\text{PQ}_c=\text{SQ}_c\cdot \text{RQ}_c$.

To provide a comprehensive evaluation, we report two versions of this metric.
First, \emph{Multi-class Panoptic Quality}~(mPQ) evaluates the model's per-class performance by calculating $\text{PQ}_c$ for each of the $K$ classes independently and then reporting the average
\begin{equation}
    \text{mPQ} = \frac{1}{K} \sum_{c=1}^K \text{PQ}_c.
\end{equation}
Second, \emph{Binary Panoptic Quality}~(bPQ) gauges the model's ability to identify instances, regardless of their type.
It is calculated by treating all classes as a single class before performing the matching and calculation.

\paragraph{Fair Evaluation of Overlapping Nuclei} \label{sec:metrics:overlaps}

A key advantage of our method is its ability to predict realistic overlapping nuclei.
However, ground-truth annotations in standard benchmarks enforce strict, often artificial separation between touching instances.
Standard PQ metrics rigidly penalize predictions that cross these boundaries, treating biological overlaps as false positives. Furthermore, attempts to resolve these overlaps through post-processing often result in incorrect instance separations that deviate from the ground truth.

To resolve this issue, we propose \emph{Masked Panoptic Quality}~(MPQ), a modification of PQ designed for this specific scenario.
MPQ uses the same core formula but replaces the standard IoU with a masked IoU (mIoU) that isolates the evaluation to the matched ground-truth instance.
Specifically, the mIoU is defined as
\begin{equation}
  \text{mIoU}(\overline m, m) = \frac{|\overline m\cap m|}{|\overline m \cup (m \setminus G)|},
\end{equation}
where $G = \bigcup_{k=1}^{M} \overline m_k$ is the union of all ground-truth masks.
Instead of including the entire area of the prediction~$m$, it considers only the area of the ground-truth mask $\overline m$ plus any part of the prediction that falls outside~$G$. By comparison, the IoU denominator is $|\overline m \cup m|$.

\paragraph{Instance Detection}

Following \citet{CellViT}, a ground-truth instance is matched to a predicted instance if the Euclidean distance between their respective centroids is less than or equal to 3~\textmu m.
Each ground-truth and predicted instance can be matched at most once.
To evaluate the model's ability to correctly localize nuclei, we use the standard detection metrics of F$_1$-score, \emph{Precision}~(P), and \emph{Recall}~(R).

\subsection{Experiments}

\subsubsection{Datasets}

For training and segmentation evaluation, we use PanNuke and MoNuSeg, both of which provide H\&E-stained RGB images paired with per-nucleus pixel-level segmentation masks. We additionally use TCGA WSIs for runtime evaluation.
These datasets provide a diverse range of samples, covering various tissue types, staining variations, and characteristics of nuclei.

\paragraph{PanNuke}

Our primary dataset for training and evaluation is PanNuke~\cite{gamper2020pannuke}.
It consists of 7,901 images, each measuring 256\texttimes256 px, collectively containing 189,744 annotated nuclei.
Crucially, these nuclei annotations are strictly non-overlapping, meaning the annotations do not model spatial overlap or occlusion between nuclei.
While the source WSIs included maximum magnifications of 20\texttimes{} and 40\texttimes{}, the dataset authors standardized and resized selected visual fields to a uniform 40\texttimes{} scale, corresponding to an effective resolution of approximately 0.25~\textmu m/px.
The dataset is highly diverse, featuring samples from 19 different tissue types and nuclei classified into five categories: neoplastic, inflammatory, dead, epithelial, and connective/soft cells.

\paragraph{MoNuSeg}

To evaluate segmentation generalization, we use the test set from the MoNuSeg dataset~\cite{8880654}.
This set contains 14 images with 7,223 annotated nuclei, sourced from seven different organs.
While the images share the same 40\texttimes{} magnification (0.25\,\textmu m/px resolution) as PanNuke, they are significantly larger (1000\texttimes1000 px).
Importantly, the nuclei in MoNuSeg are not categorized by cell type.
Unlike PanNuke's strictly non-overlapping masks, some instances in the released MoNuSeg dataset share small overlapping regions. These overlaps are artifacts of the manual delineation process rather than intentional modeling of occlusion between nuclei.
To maintain consistency with other methods, our evaluation uses the official challenge labels, in which these overlap pixels are assigned to the nucleus appearing on top to produce mutually exclusive masks.

\paragraph{TCGA}

We evaluate the computational performance of our model on WSIs selected from the TCGA-COAD/READ cohort.
To ensure a direct comparison, we used the same set of images as the HoVer-NeXt benchmark~\cite{baumann2024hover}.
This set includes images with tissue sizes ranging from 20 mm\textsuperscript{2} to 500 mm\textsuperscript{2} (estimated using their algorithm), with further details provided in~\cref{tab:wsi}.
\begin{table}[t]
    \centering
    \small
    \caption{Representative images from the TCGA-COAD/READ cohort used for inference-speed evaluation. To ensure a direct comparison, we used the same set of images and the same tissue-area estimation as~\citet{baumann2024hover}.
    }
    \begin{tabular*}{\linewidth}{@{}l@{\;\extracolsep{\fill}}c@{\;}rcr@{}}
        \toprule
        Case ID & Slide ID & H\texttimes W (px) & \textmu m/px & mm\textsuperscript{2}\\
        \midrule
        TCGA-AA-3977 & DX1 &  57,600\texttimes  34,560 & 0.2325 &  20.16  \\
        TCGA-AA-3688 & DX1 & 101,120\texttimes  66,816 & 0.2325 &  49.84  \\
        TCGA-AA-A010 & DX1 &  68,352\texttimes  71,424 & 0.2325 & 101.09  \\
        TCGA-CK-4951 & DX1 &  64,833\texttimes  94,965 & 0.2520 & 202.90  \\
        TCGA-5M-AAT5 & DX1 &  89,145\texttimes 161,352 & 0.2525 & 501.00  \\
        \bottomrule
    \end{tabular*}
    \label{tab:wsi}
\end{table}

\subsubsection{Experimental Setup}

\paragraph{Input Preprocessing}

To standardize the input data, we normalize the RGB channels of each image using a mean $\mu = [0.485, 0.456, 0.406]$ and a standard deviation $\sigma = [0.229, 0.224, 0.225]$, consistent with the ImageNet pretraining of the backbone.

To efficiently determine the ground-truth radial distance bounds ($r_\text{min}$ and $r_\text{max}$) for any point $p$ along a specific ray, our framework uses a dynamic, on-the-fly lookup strategy during data loading. For each training sample, we construct two localized lookup tables (one for $r_\text{min}$ and one for $r_\text{max}$) using a custom algorithm inspired by StarDist~\cite{StarDist}. Specifically, for each pixel within the $R \times R$ image space, the algorithm computes the distance to the object boundary along 64 fixed radial directions. The resulting minimum and maximum boundary distances for each of the 64 rays are stored in their respective tables.
Crucially, this computation is offloaded to the CPU and executed in parallel during the data sampling stage, ensuring that the GPU remains fully utilized for network training without blocking overhead. Retrieving the bounds for any matched foreground query reduces to a highly efficient $O(1)$ GPU tensor lookup. For points that fall between discrete pixel coordinates, bilinear interpolation is applied across the spatial dimensions to ensure a smooth, continuous distance field. This approach maintains a manageable memory footprint of $2 \times 64 \times R \times R$ float32 values per sample while bypassing GPU execution bottlenecks.

\paragraph{Hyperparameter Settings}
Following DETR, we use a post-layer normalization transformer.
The model is configured with a token dimension of $d=384$, $L=6$ transformer layers with 12 attention heads, and a SwiGLU-based feed-forward network (FFN\textsubscript{SwiGLU})~\cite{shazeer2020glu} with an intermediate dimension of~1024.

We employ STA~\cite{zhang2025fast} for all attention layers.
For self-attention, we use a 3\texttimes3 tile size for object queries with an attention window of 3 to facilitate deduplication.
For cross-attention, the queries retain their 3\texttimes3 tile size, but we widen the attention window to 5 to better capture large nuclei from the multi-scale image feature maps, which are tiled at 2\texttimes2, 4\texttimes4, and 8\texttimes8 for resolutions 1/16, 1/8, and 1/4, respectively.
Finally, we use RoPE with learnable frequencies initialized with a base of $\theta=100$.

\paragraph{Training}
To ensure a fair comparison, our training regimen closely follows that of CellViT~\cite{CellViT}.
Models are trained for 130 epochs with a batch size of 16, following the staged approach described in that study: the backbone weights are frozen for the first 30 epochs and subsequently fine-tuned with a learning rate set to $10\,\%$ of the main learning rate.
Furthermore, we use the same data augmentation and sampling strategies described in that~study.
\looseness=-1

For optimization, we use the AdamW optimizer with an initial learning rate of $10^{-4}$ and a weight decay of $10^{-4}$.
The learning rate is controlled using a cosine annealing schedule with a 10-epoch linear warm-up.
To improve efficiency, we employ float16 mixed precision and apply gradient norm clipping with a threshold of 0.1.
FLOPs are measured on a single 256\texttimes256 input image with a resolution of 0.25~\textmu m/px.

\paragraph{Baseline Methods}
To ensure a fair and comprehensive comparison, we used the highest-performing variants of state-of-the-art methods.
For CellViT~\cite{CellViT}, we employed the CellViT-SAM-H version.
We evaluated LKCell~\cite{LKCell} using its large variant, LKCell-L.
For HoVer-NeXt~\cite{baumann2024hover}, we selected the HN$_{Tiny, 16TTA}$ configuration, as it offers a strong balance of speed and accuracy.
Finally, CPP-Net~\cite{chen2023cpp} was evaluated with a ResNet50~\cite{resnet} backbone.

\paragraph{Inference and Evaluation}
We apply distance-transform watershed refinement to our predictions when computing panoptic quality (bPQ, mPQ) on the PanNuke and MoNuSeg datasets.
This is required for two reasons: PanNuke annotations are strictly non-overlapping, and the standard Panoptic Quality (PQ) metric forbids overlapping predictions.
This refinement resolves overlaps by assigning shared pixels to the instance with the largest Euclidean distance from its boundary. For all other evaluation metrics, we use the raw, unaltered predictions.

To obtain direct evidence for the biological plausibility of predicted overlaps among ambiguous neighboring nuclei, we evaluated three trained PanNuke models on their held-out test folds, selecting pairs of touching ground-truth annotations that matched model predictions ($\text{IoU} > 0.5$). These pairs were stratified into \say{overlapping} and \say{touching} strata based on the intersection area of their corresponding model-predicted polygons, using an initial threshold of $0.56 \text{ \textmu m}^2$ ($9 \text{ px}^2$). From these strata, we randomly sampled 100 pairs per stratum for each of the three models (600 pairs in total) for an initial review by an expert pathologist, who classified each as \say{touching}, \say{overlapping}, or \say{unsure} using only the nuclei centroids to prevent geometric bias. To rigorously validate the models, we then drew a stratified sample of 210 pairs (35 per model per stratum) from this initial pool for an independent review by a second pathologist. To formally quantify inter-rater reliability between the two experts, we calculated Cohen's kappa on this 210-pair subset, thereby accounting for chance agreement. The remaining 390 pairs, excluding any \say{unsure} pairs, served as a tuning set for identifying the optimal area threshold for the intersection of the predicted polygons that maximized the F$_1$-score. Establishing this threshold empirically on a tuning set was necessary because the precise, implicit visual cutoff that pathologists use to mentally distinguish between a \say{touch} and a true \say{overlap} is unknown and highly subjective. We subsequently applied this optimal threshold to the consensus pairs within the 210-pair test set--specifically, those for which both pathologists agreed on the classification--to compute precision, recall, F$_1$-score, specificity, and AUROC. All performance estimates incorporated inverse-probability sampling weights (IPW) to correct for the disproportionate stratified sampling. Finally, statistical uncertainty was assessed using 10,000 nonparametric bootstrap replicates, with stratified resampling within the original test fold and sampling stratum while the previously identified optimal threshold was kept fixed.

Except for CellViT, the inference-speed benchmarks were conducted on a system equipped with 12 CPU cores, 64 GB of RAM, and an NVIDIA H100 NVL GPU. Before recording timings, we performed a warm-up run using the smallest whole-slide image and excluded it from the measurements. For the locally executed pipelines other than HoVer-NeXt, elapsed time included tiling, tissue-mask calculation, model inference, method-specific post-processing, and output saving. These measurements approximate a practical WSI-processing workload rather than a complete clinical workflow. For LKCell, StarDist, and CPP-Net (sourced from CellSeg Models \cite{cellseg_models}) and for LSP-DETR, we used our optimized multiprocessing pipeline. We used a tile size of 2048\texttimes 2048 with a batch size of 20, which we found optimal for our hardware and data-pipeline architecture (e.g., disk I/O speed). The StarDist implementation does not support batch inference, but we adapted its workflow. All four pipelines used a 16-pixel inter-tile overlap to maximize comparability with the HoVer-NeXt study. HoVer-NeXt was run using its official implementation with the HN$_{\text{Tiny,16TTA}}$ configuration, 256\texttimes 256 tiles, a 16-pixel overlap, and a batch size of 64. For HoVer-NeXt, we omitted its post-processing because this cost was amortized through parallelism for the other methods in our pipeline. Prediction aggregation and final output saving were excluded; its reported runtime is therefore not a complete end-to-end measurement. The CellViT runtime was estimated from Baumann et al.'s benchmark~\cite{baumann2024hover}, which used an NVIDIA A40 and 16 CPU cores under a different protocol. The resulting value is a cross-hardware extrapolation and not comparable to the local H100 measurements.

\section{Results}

\subsection{Evaluation on the PanNuke Dataset}

\begin{table*}[t]
    \centering
    \caption{
        Performance of instance segmentation on the PanNuke dataset across the 19 tissue types using three-fold cross-validation. The standard deviation (STD) across the three splits is provided in the final row.
        The best bPQ and mPQ results are highlighted.
        StarDist was trained by~\citet{chen2023cpp}.}
    \begin{tabular}{lcgcgcgcgcgcg}
        \toprule
        
         \multirow{2}{*}[-2pt]{Tissue} & \multicolumn{2}{c}{\textbf{HoVer-NeXt}} & \multicolumn{2}{c}{\textbf{StarDist}} & \multicolumn{2}{c}{\textbf{CPP-Net}} & \multicolumn{2}{c}{\textbf{CellViT}} & \multicolumn{2}{c}{\textbf{LKCell}} & \multicolumn{2}{c}{\textbf{LSP-DETR}} \\

        \cmidrule(lr){2-3} \cmidrule(lr){4-5} \cmidrule(lr){6-7} \cmidrule(lr){8-9} \cmidrule(lr){10-11} \cmidrule(lr){12-13}

        & mPQ & \wb bPQ & mPQ & \wb bPQ & mPQ & \wb bPQ & mPQ & \wb bPQ & mPQ & \wb bPQ & mPQ & \wb bPQ \\
        
        \midrule

        Adrenal    & 49.4      & 70.4 & 48.7      & 69.7 & 49.4 & 70.7      & \bf 51.3      & 70.9 & 50.8 & \bf 71.3 & 47.1 & 70.9 \\
        Bile Duct  & 46.8      & 66.8 & 46.5      & 66.9 & 46.7 & 67.7      & \bf 48.9      & 67.8 & \bf 48.9      & \bf 68.2      & 46.3 & 67.2 \\
        Bladder    & 57.8      & 69.6 & 57.9      & 69.9 & 59.4 & 70.5      & 58.4      & 70.7 & \bf 59.9      & \bf 71.5 & 58.3 & 71.1\\
        Breast     & 49.5      & 64.3 & 50.6      & 66.7 & 50.9 & 67.5      & \bf 51.8      & 67.5 & 51.4      & 66.9      & 51.4 & \bf 67.6 \\
        Cervix     & 47.5      & 66.7 & 46.3      & 66.9 & 47.9 & 69.1      & 49.8      & 68.7 & \bf 49.9      & \bf 69.6      & 49.3 & 68.8 \\
        Colon      & 42.8      & 57.0 & 42.1      & 57.8 & 43.2 & 59.1      & 44.9      & 59.2 & \bf 45.6      & 59.4      & 44.4 & \bf 59.6 \\
        Esophagus  & 52.7      & 64.7 & 53.3      & 66.6 & 54.5 & 68.0      & 54.5      & 66.8 & \bf 56.1      & \bf 68.3 & 54.3 & 67.7 \\
        Head\&Neck & 48.5      & 64.3 & 47.7      & 64.3 & 47.1 & 65.2      & 49.1      & 65.4 & \bf 50.0      & \bf 66.1      & 48.4 & 64.8 \\
        Kidney     & 51.7      & 68.3 & 48.8      & 70.0 & 51.9 & 70.7      & 53.7      & 70.9 & \bf 55.8      & \bf 72.0      & 52.4 & 69.9 \\
        Liver      & 50.4      & 71.7 & 51.5      & 72.3 & 51.4 & 73.1      & 52.2      & 73.2 & \bf 53.0      & \bf 73.7      & 49.8 & 72.1 \\
        Lung       & 42.9      & 63.4 & 41.3      & 63.6 & 42.6 & 63.9      & \bf 43.1      & 64.3 & \bf 43.1      & \bf 64.9 & 42.9 & 64.0 \\
        Ovarian    & 48.6      & 61.2 & 52.1      & 66.7 & 53.1 & \bf 68.3      & 53.9      & 67.2 & \bf 54.4      & 67.6 & 51.6 & 67.6 \\
        Pancreatic & 46.0      & 65.7 & 45.9      & 66.0 & 47.1 & \bf 67.9      & 47.2      & 66.6 & \bf 49.0      & 67.7 & 47.9 & 67.1 \\
        Prostate   & 48.1      & 62.9 & 50.7      & 67.5 & 53.1 & \bf 69.0 & \bf 53.2      & 68.2 & 52.3      & 68.1      & 51.6 & 67.5 \\
        Skin       & 41.4      & 62.3 & 36.1      & 62.9 & 35.7 & 62.1      & \bf 43.4      & 65.7 & 40.8      & \bf 66.0      & 37.7 & 64.1 \\
        Stomach    & 46.1      & 69.6 & 44.8      & 69.4 & 45.8 & 70.7      & \bf 47.1      & 70.2 & 46.6      & \bf 71.5 & 43.7 & 68.7 \\
        Testis     & 49.7      & 68.0 & 49.4      & 68.7 & 49.3 & \bf 70.3      & 51.3      & 69.6 & \bf 53.5      & 69.4 & 48.7 & 68.7 \\
        Thyroid    & 42.2      & 67.7 & 43.0      & 69.6 & 43.9 & \bf 71.6    & 45.2      & 71.5 & \bf 46.6      & 71.5 & 44.8 & 70.3 \\
        Uterus     & 44.6      & 61.9 & 44.8      & 66.0 & 47.9 & 66.2      & 47.4      & 66.3 & \bf 49.0      & \bf 66.6 & 45.9 & 64.8 \\
        \midrule
        Average    & 47.7      & 65.6 & 47.4      & 66.9 & 48.5 & 68.0      & 49.8      & 67.9 & \bf 50.3      & \bf 68.4 & 48.2 & 67.5 \\

        STD & -- & -- & -- & -- & -- & -- & 0.77 & 0.39 & 0.96 & 0.25 & 0.83 & 0.24 \\
        \bottomrule
    \end{tabular}
    \label{tab:pannuke}
\end{table*}

We follow the standard three-fold cross-validation described in~\cite{graham2019hover}.
For evaluation, we use the final training checkpoint from each of the three folds to avoid selection bias, and all reported metrics represent the average performance across the three folds.

\Cref{tab:pannuke} provides a comprehensive comparison of panoptic quality across 19 tissue types using bPQ and mPQ metrics.
Notably, the LKCell metrics reported in these tables were recomputed from official result logs to resolve discrepancies in the original publication and ensure a fair comparison.

We also report MPQ on the raw predictions to evaluate overlap-capable outputs without imposing watershed-based separation.
This metric effectively softens the boundaries between touching ground-truth nuclei, allowing us to assess performance without the unfair penalty imposed on realistic overlaps.
LSP-DETR achieves an average mMPQ of 49.0 and a bMPQ of 68.8.
While we cannot re-evaluate other methods using MPQ due to the lack of raw predictions or pretrained checkpoints, MPQ is, by construction, greater than or equal to PQ (see \cref{sec:metrics:overlaps}). The magnitude of the increase cannot be determined without the predictions; however, across our evaluations, it was consistently below one point for methods producing non-overlapping predictions. MPQ differs from standard PQ only in regions containing densely annotated nuclei, where its relaxed treatment of touching boundaries can increase the matching quality.
We further analyze performance across PanNuke's five distinct cell categories.
\Cref{tab:pannuke:classification} compares panoptic quality for each cell type using the aforementioned watershed refinement, while~\cref{tab:pannuke:classification:detection} isolates detection performance by reporting detection-quality metrics for each cell category.
\begin{table}[t]
    \centering
    \footnotesize
    \caption{Comparison of Panoptic Quality (PQ) for each cell category on the PanNuke dataset. The PQ scores are averaged across the three folds of cross-validation.}
    \begingroup
    \setlength{\tabcolsep}{2pt}
    \begin{tabular*}{\linewidth}{@{}l@{\extracolsep{\fill}}ccccc@{}}
        \toprule
        Model & Neoplastic & Inflammatory & Dead & Connective & Epithelial \\
        \midrule
        HoVer-NeXt & 53.6 & 41.8 & 15.4 & 41.5 & 51.5 \\
        CellViT    & 58.1 & 41.7 & 14.9 & 42.3 & 58.3 \\
        LKCell     & \bf 58.4 & \bf 43.7 & \bf 15.5 & \bf 42.5 & \bf 58.5 \\
        LSP-DETR   & 57.4 & 41.3 & 13.0 & 40.8 & 54.5 \\
        \bottomrule
    \end{tabular*}
    \endgroup
    \label{tab:pannuke:classification}
    \vspace{-1ex}
\end{table}

\begin{table*}[t]
    \centering
    \small
    \caption{Detection performance across individual PanNuke categories. The scores are averaged across the three folds of cross-validation.}
    \resizebox{\linewidth}{!}{
    \begin{tabular}{lccc|ccc|ccc|ccc|ccc}
        \toprule

        \multicolumn{1}{c}{\multirow{2}{*}{Model}} & \multicolumn{3}{c}{\textbf{Neoplastic}} & \multicolumn{3}{c}{\textbf{Inflammatory}} & \multicolumn{3}{c}{\textbf{Dead}} & \multicolumn{3}{c}{\textbf{Connective}} & \multicolumn{3}{c}{\textbf{Epithelial}} \\

        \cmidrule(lr){2-4} \cmidrule(lr){5-7} \cmidrule(lr){8-10} \cmidrule(lr){11-13} \cmidrule(lr){14-16}

        \multicolumn{1}{c}{} & \multicolumn{1}{c}{P} & \multicolumn{1}{c}{R} & \multicolumn{1}{c}{$\text{F}_1$} & \multicolumn{1}{c}{P} & \multicolumn{1}{c}{R} & \multicolumn{1}{c}{$\text{F}_1$} & \multicolumn{1}{c}{P} & \multicolumn{1}{c}{R} & \multicolumn{1}{c}{$\text{F}_1$} & \multicolumn{1}{c}{P} & \multicolumn{1}{c}{R} & \multicolumn{1}{c}{$\text{F}_1$} & P & R & $\text{F}_1$ \\
        
        \midrule

        HoVer-NeXt & - & - & 72.0 & - & - & 68.1 & - & - & \bf 49.2 & - & - & \bf 64.6 & - & - & 72.8 \\
        CellViT & 71.8 & 69.3 & 70.5 & 59.3 & 57.4 & 58.3 & 43.0 & 31.6 & 36.3 & 54.5 & 52.5 & 53.5 & 72.4 & 73.2 & 72.8 \\
        LKCell & 70.3 & 71.7 & 71.0 & 58.2 & 59.9 & 59.0 & 45.3 & 34.5 & 39.1 & 57.3 & 50.2 & 53.5 & 74.0 & \bf 75.5 & 74.7 \\
        LSP-DETR & \bf 76.9 & \bf 73.2 & \bf 75.0 & \bf 71.1 & \bf 68.4 & \bf 69.7 & \bf 59.4 & \bf 39.8 & 47.4 & \bf 66.8 & \bf 59.1 & 62.7 & \bf 77.2 & 73.6 & \bf 75.4 \\

        \bottomrule
    \end{tabular}
    }
    \label{tab:pannuke:classification:detection}
\end{table*}

\subsection{Pathologist Evaluation of Predicted Overlaps}

\begin{table}[!t]
    \centering
    \small
    \caption{Inter-observer agreement between two pathologists evaluating a total of 210 nucleus pairs. These were obtained by randomly selecting exactly 35 samples from each of the six test-set strata.}
    \begin{tabular*}{\linewidth}{@{}l@{\extracolsep{\fill}}ccc@{}}
        \toprule
        Pat. 1 \textbackslash{} Pat. 2 & Overlapping & Touching & Unsure \\
        \midrule
        Overlapping & 78 & 1  & 23 \\
        Touching    & 27 & 37 & 6  \\
        Unsure      & 12 & 2  & 24 \\
        \bottomrule
    \end{tabular*}
    \label{tab:pathologist_agreement}
\end{table}

Inter-observer agreement between the two expert pathologists yielded a Cohen's kappa of 0.46 (\cref{tab:pathologist_agreement}). The pathologists reached consensus on 115 pairs (78 overlapping and 37 touching). On this consensus set, the continuous polygon-overlap score achieved an IPW-weighted AUROC of 0.983 (95\,\% CI, 0.969--0.994). At the pre-established operating threshold of $0.16 \text{ \textmu m}^2$, the models yielded the following IPW-weighted performance metrics: F$_1$-score, 0.964 (95\,\% CI, 0.946--0.980); precision, 0.977 (95\,\% CI, 0.955--0.994); recall, 0.950 (95\,\% CI, 0.920--0.978); and specificity, 0.892 (95\,\% CI, 0.785--0.975).

\subsection{Evaluation on the MoNuSeg Dataset}

To assess generalization and scalability, we evaluated our PanNuke-trained models (from all three cross-validation folds) directly on the MoNuSeg test set.
This evaluation highlights our model's input-size agnosticism: despite being trained exclusively on 256\texttimes 256 px patches, our method processes each 1000\texttimes 1000 px MoNuSeg image in its entirety at inference.
Each image is processed in a single pass, eliminating the need for the overlapping patch-based strategies required by many other methods.

The results are reported in~\cref{tab:monuseg}.
As MoNuSeg does not provide cell-type labels, cell-category predictions were ignored during this evaluation.

\subsection{Computational Performance Evaluation on the TCGA Dataset}
The most significant advantage of LSP-DETR is its computational efficiency, as summarized in \cref{tab:comparison}.
Unlike the other evaluated methods, LSP-DETR is a fully end-to-end model that completely eliminates the need for expensive post-processing.
On the TCGA dataset, LSP-DETR achieves an inference speed of 0.45~s/mm\textsuperscript{2}, more than 3\texttimes{} faster than the next-fastest, StarDist (1.42~s/mm\textsuperscript{2}), and 5\texttimes{} faster than HoVer-NeXt (2.38~s/mm\textsuperscript{2}).

\vspace{1.7ex}
\section{Discussion}

Our results demonstrate that LSP-DETR is a highly effective and computationally efficient method for nuclei segmentation.
Evaluation across the PanNuke, MoNuSeg, and TCGA datasets highlights four core advantages: robust segmentation and detection performance, strong generalization, inference speed that is strictly better than the state of the art, and the ability to model realistic overlaps between nuclei.

\begin{figure*}[t!]
    \centering
    \setlength{\tabcolsep}{0pt} 
    \renewcommand{\arraystretch}{0} 
    
    \newlength{\imgwidth} 
    \setlength{\imgwidth}{0.112\textwidth} 
    
    \newcommand{\hpanelgap}{0.8em} 
    \newcommand{\vpanelgap}{0em}   
    \newcommand{\imagegap}{2pt}    

    \newcommand{\swatch}[1]{\fcolorbox{black}{#1}{\rule{0pt}{1em}\rule{2em}{0pt}}}

    \begin{tabular}{c@{\hspace{1ex}} c@{\hspace{\imagegap}}c@{\hspace{\hpanelgap}} c@{\hspace{\imagegap}}c@{\hspace{\hpanelgap}} c@{\hspace{\imagegap}}c@{\hspace{\hpanelgap}} c@{\hspace{\imagegap}}c}
        
        & \multicolumn{2}{c@{\hspace{\hpanelgap}}}{\rule{0pt}{2ex}\scriptsize Adrenal} 
        & \multicolumn{2}{c@{\hspace{\hpanelgap}}}{\rule{0pt}{2ex}\scriptsize Bile duct} 
        & \multicolumn{2}{c@{\hspace{\hpanelgap}}}{\rule{0pt}{2ex}\scriptsize Bladder} 
        & \multicolumn{2}{c}{\rule{0pt}{2ex}\scriptsize Breast} \\[0.5ex] 
        
        \raisebox{\dimexpr 0.5\imgwidth - 0.5\height\relax}[0pt][0pt]{\rotatebox{90}{\scriptsize GT}} &
        \includegraphics[width=\imgwidth]{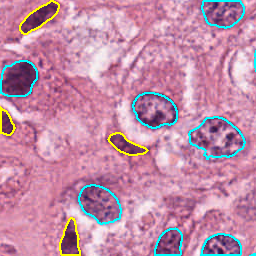} &
        \includegraphics[width=\imgwidth]{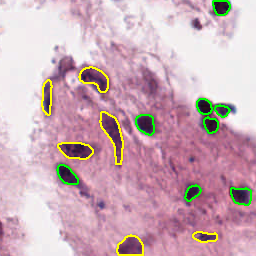} &
        \includegraphics[width=\imgwidth]{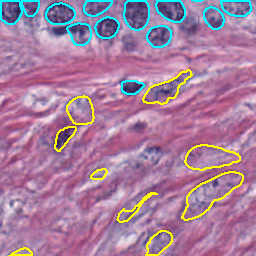} &
        \includegraphics[width=\imgwidth]{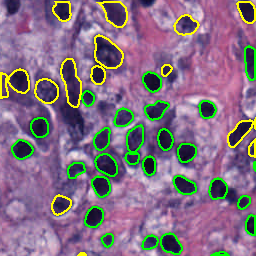} &
        \includegraphics[width=\imgwidth]{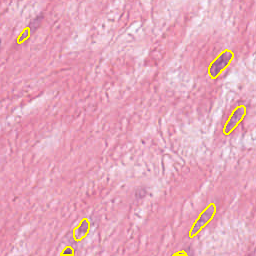} &
        \includegraphics[width=\imgwidth]{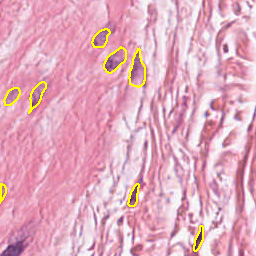} &
        \includegraphics[width=\imgwidth]{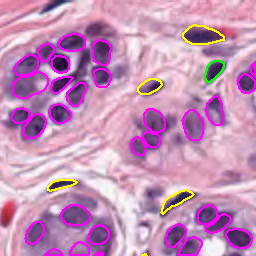} &
        \includegraphics[width=\imgwidth]{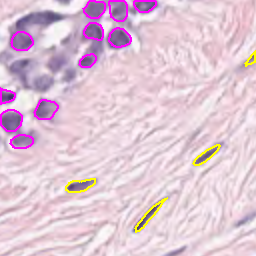} \\[\imagegap] 
        
        \raisebox{\dimexpr 0.5\imgwidth - 0.5\height\relax}[0pt][0pt]{\rotatebox{90}{\scriptsize Prediction}} &
        \includegraphics[width=\imgwidth]{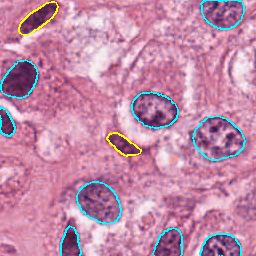} &
        \includegraphics[width=\imgwidth]{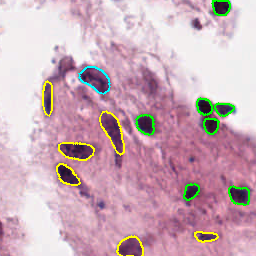} &
        \includegraphics[width=\imgwidth]{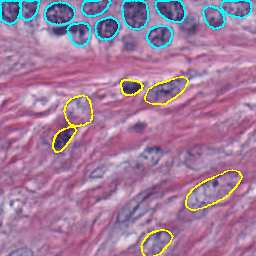} &
        \includegraphics[width=\imgwidth]{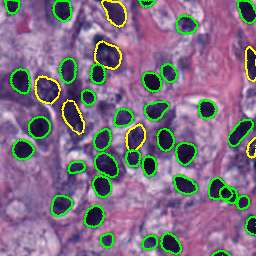} &
        \includegraphics[width=\imgwidth]{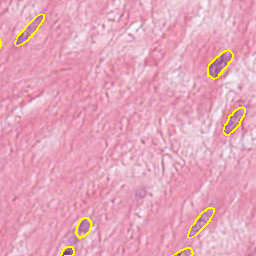} &
        \includegraphics[width=\imgwidth]{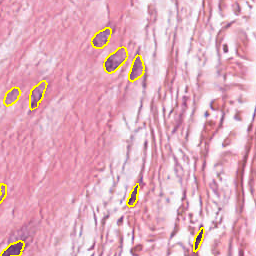} &
        \includegraphics[width=\imgwidth]{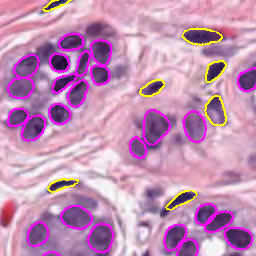} &
        \includegraphics[width=\imgwidth]{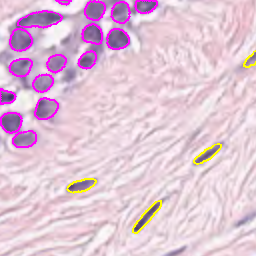} \\[\vpanelgap]

        & \multicolumn{2}{c@{\hspace{\hpanelgap}}}{\rule{0pt}{2ex}\scriptsize Cervix} 
        & \multicolumn{2}{c@{\hspace{\hpanelgap}}}{\rule{0pt}{2ex}\scriptsize Colon} 
        & \multicolumn{2}{c@{\hspace{\hpanelgap}}}{\rule{0pt}{2ex}\scriptsize Esophagus} 
        & \multicolumn{2}{c}{\rule{0pt}{2ex}\scriptsize Head \& Neck} \\[0.5ex] 
        
        \raisebox{\dimexpr 0.5\imgwidth - 0.5\height\relax}[0pt][0pt]{\rotatebox{90}{\scriptsize GT}} &
        \includegraphics[width=\imgwidth]{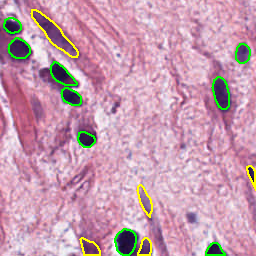} &
        \includegraphics[width=\imgwidth]{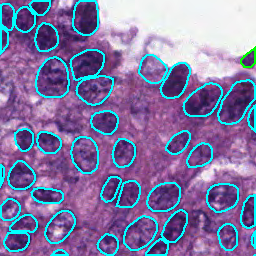} &
        \includegraphics[width=\imgwidth]{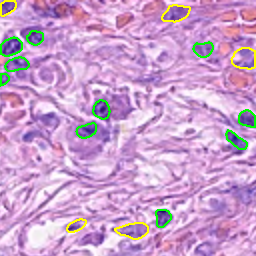} &
        \includegraphics[width=\imgwidth]{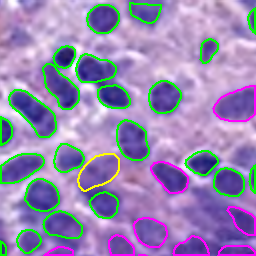} &
        \includegraphics[width=\imgwidth]{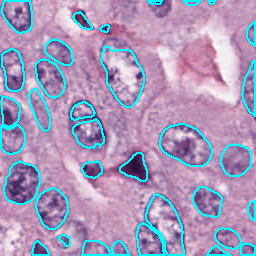} &
        \includegraphics[width=\imgwidth]{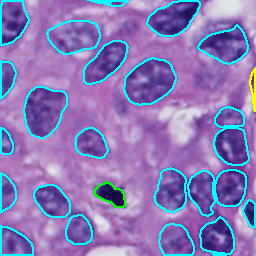} &
        \includegraphics[width=\imgwidth]{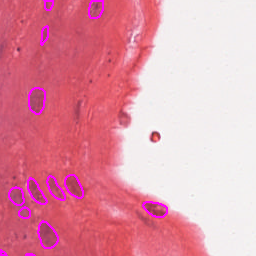} &
        \includegraphics[width=\imgwidth]{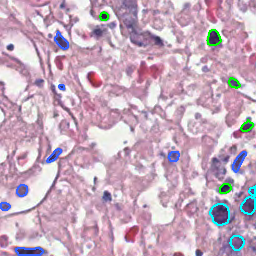} \\[\imagegap] 
        
        \raisebox{\dimexpr 0.5\imgwidth - 0.5\height\relax}[0pt][0pt]{\rotatebox{90}{\scriptsize Prediction}} &
        \includegraphics[width=\imgwidth]{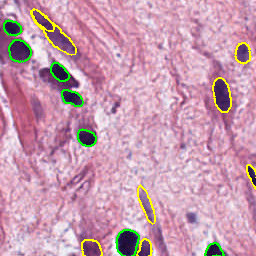} &
        \includegraphics[width=\imgwidth]{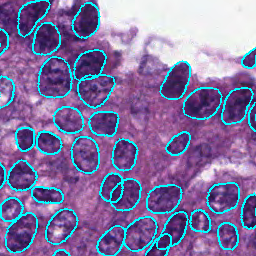} &
        \includegraphics[width=\imgwidth]{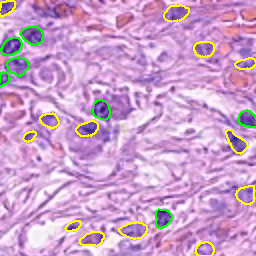} &
        \includegraphics[width=\imgwidth]{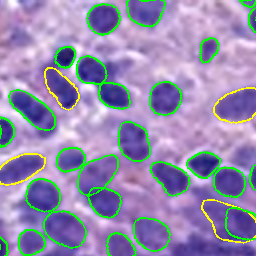} &
        \includegraphics[width=\imgwidth]{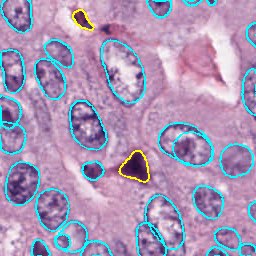} &
        \includegraphics[width=\imgwidth]{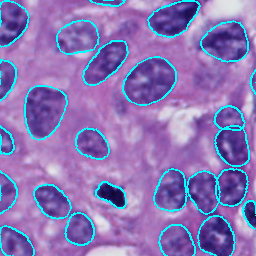} &
        \includegraphics[width=\imgwidth]{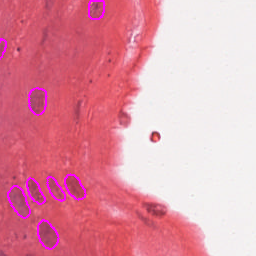} &
        \includegraphics[width=\imgwidth]{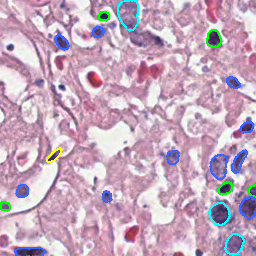} \\[\vpanelgap]

        & \multicolumn{2}{c@{\hspace{\hpanelgap}}}{\rule{0pt}{2ex}\scriptsize Kidney} 
        & \multicolumn{2}{c@{\hspace{\hpanelgap}}}{\rule{0pt}{2ex}\scriptsize Liver} 
        & \multicolumn{2}{c@{\hspace{\hpanelgap}}}{\rule{0pt}{2ex}\scriptsize Lung} 
        & \multicolumn{2}{c}{\rule{0pt}{2ex}\scriptsize Ovarian} \\[0.5ex] 
        
        \raisebox{\dimexpr 0.5\imgwidth - 0.5\height\relax}[0pt][0pt]{\rotatebox{90}{\scriptsize GT}} &
        \includegraphics[width=\imgwidth]{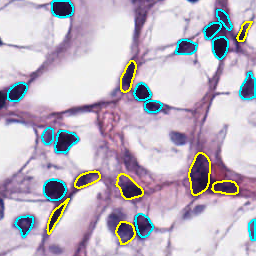} &
        \includegraphics[width=\imgwidth]{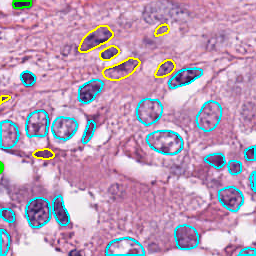} &
        \includegraphics[width=\imgwidth]{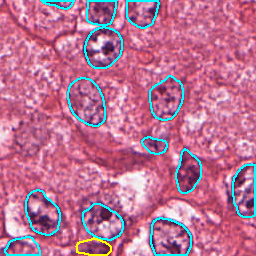} &
        \includegraphics[width=\imgwidth]{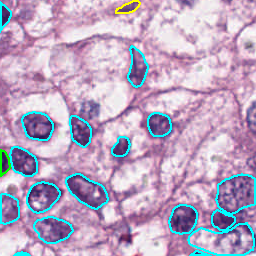} &
        \includegraphics[width=\imgwidth]{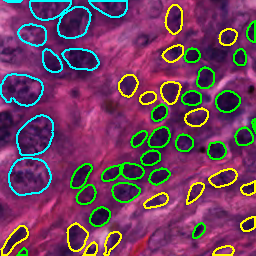} &
        \includegraphics[width=\imgwidth]{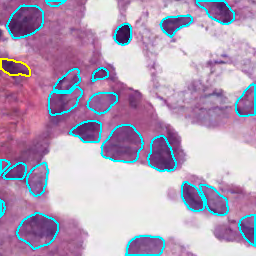} &
        \includegraphics[width=\imgwidth]{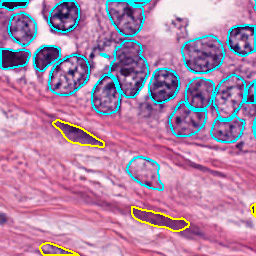} &
        \includegraphics[width=\imgwidth]{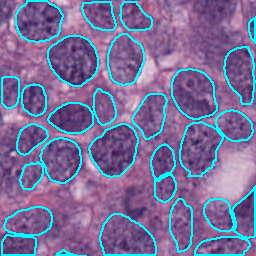} \\[\imagegap] 
        
        \raisebox{\dimexpr 0.5\imgwidth - 0.5\height\relax}[0pt][0pt]{\rotatebox{90}{\scriptsize Prediction}} &
        \includegraphics[width=\imgwidth]{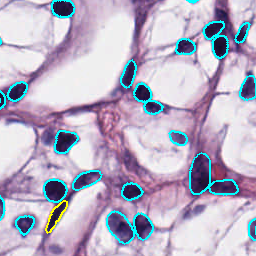} &
        \includegraphics[width=\imgwidth]{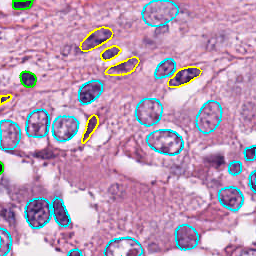} &
        \includegraphics[width=\imgwidth]{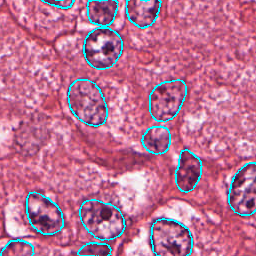} &
        \includegraphics[width=\imgwidth]{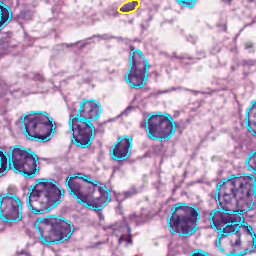} &
        \includegraphics[width=\imgwidth]{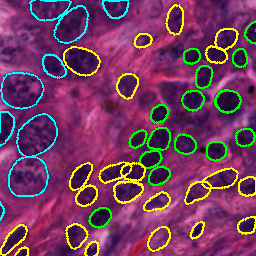} &
        \includegraphics[width=\imgwidth]{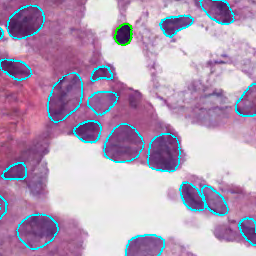} &
        \includegraphics[width=\imgwidth]{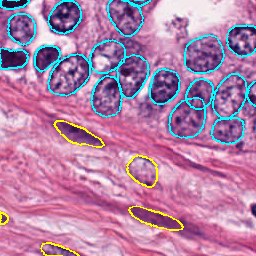} &
        \includegraphics[width=\imgwidth]{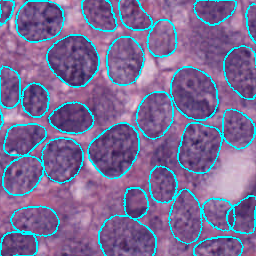} \\[\vpanelgap]

        & \multicolumn{2}{c@{\hspace{\hpanelgap}}}{\rule{0pt}{2ex}\scriptsize Pancreatic} 
        & \multicolumn{2}{c@{\hspace{\hpanelgap}}}{\rule{0pt}{2ex}\scriptsize Prostate} 
        & \multicolumn{2}{c@{\hspace{\hpanelgap}}}{\rule{0pt}{2ex}\scriptsize Skin} 
        & \multicolumn{2}{c}{\rule{0pt}{2ex}\scriptsize Stomach} \\[0.5ex] 
        
        \raisebox{\dimexpr 0.5\imgwidth - 0.5\height\relax}[0pt][0pt]{\rotatebox{90}{\scriptsize GT}} &
        \includegraphics[width=\imgwidth]{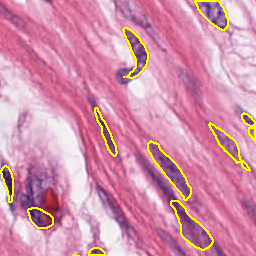} &
        \includegraphics[width=\imgwidth]{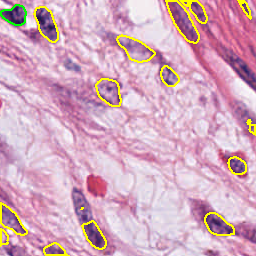} &
        \includegraphics[width=\imgwidth]{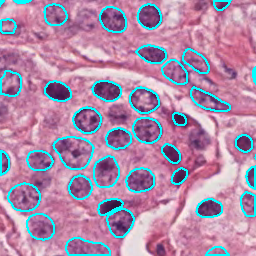} &
        \includegraphics[width=\imgwidth]{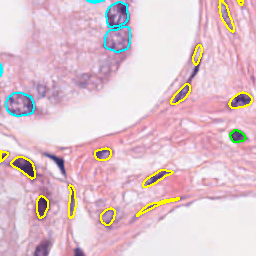} &
        \includegraphics[width=\imgwidth]{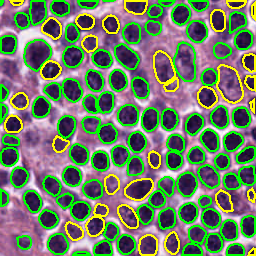} &
        \includegraphics[width=\imgwidth]{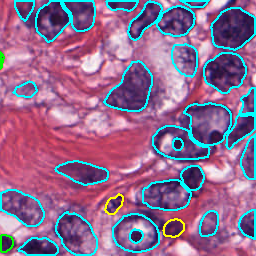} &
        \includegraphics[width=\imgwidth]{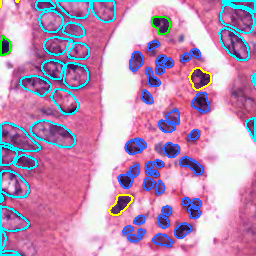} &
        \includegraphics[width=\imgwidth]{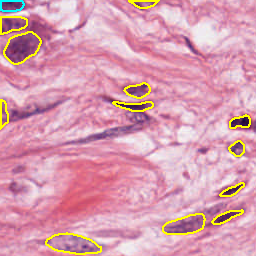} \\[\imagegap] 
        
        \raisebox{\dimexpr 0.5\imgwidth - 0.5\height\relax}[0pt][0pt]{\rotatebox{90}{\scriptsize Prediction}} &
        \includegraphics[width=\imgwidth]{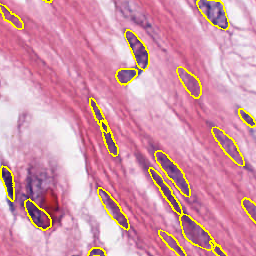} &
        \includegraphics[width=\imgwidth]{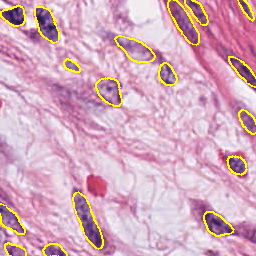} &
        \includegraphics[width=\imgwidth]{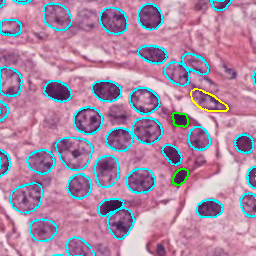} &
        \includegraphics[width=\imgwidth]{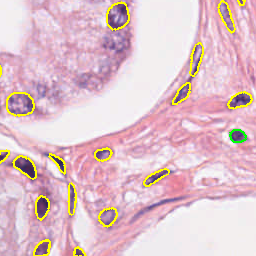} &
        \includegraphics[width=\imgwidth]{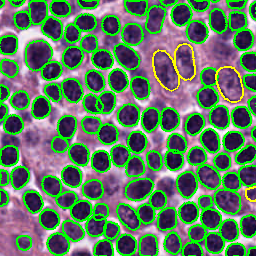} &
        \includegraphics[width=\imgwidth]{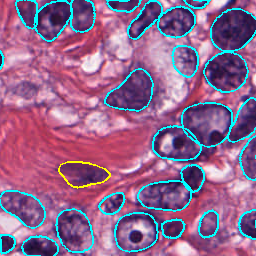} &
        \includegraphics[width=\imgwidth]{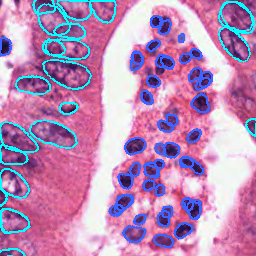} &
        \includegraphics[width=\imgwidth]{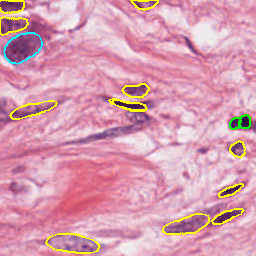} \\[\vpanelgap]

        & \multicolumn{2}{c@{\hspace{\hpanelgap}}}{\rule{0pt}{2ex}\scriptsize Testis} 
        & \multicolumn{2}{c@{\hspace{\hpanelgap}}}{\rule{0pt}{2ex}\scriptsize Thyroid} 
        & \multicolumn{2}{c@{\hspace{\hpanelgap}}}{\rule{0pt}{2ex}\scriptsize Uterus} 
        & \multicolumn{2}{c}{} \\[0.5ex] 
        
        \raisebox{\dimexpr 0.5\imgwidth - 0.5\height\relax}[0pt][0pt]{\rotatebox{90}{\scriptsize GT}} &
        \includegraphics[width=\imgwidth]{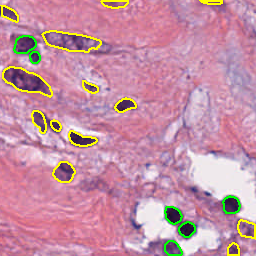} &
        \includegraphics[width=\imgwidth]{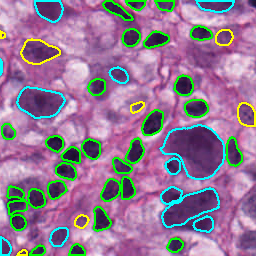} &
        \includegraphics[width=\imgwidth]{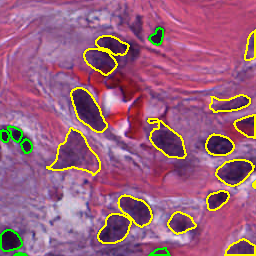} &
        \includegraphics[width=\imgwidth]{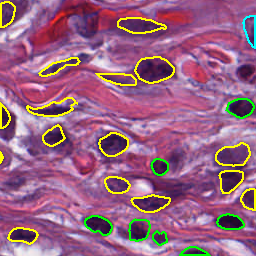} &
        \includegraphics[width=\imgwidth]{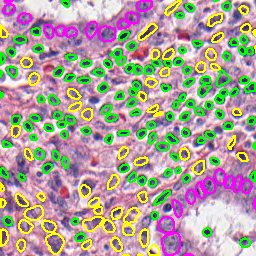} &
        \includegraphics[width=\imgwidth]{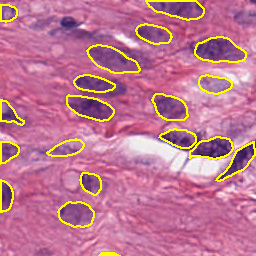} &
        \multicolumn{2}{c}{
            \raisebox{\dimexpr -0.4\imagegap - 0.3\height\relax}[0pt][0pt]{
                {
                \setlength{\fboxsep}{0pt}\setlength{\fboxrule}{0.1pt}%
                \renewcommand{\arraystretch}{0.9}
                \scriptsize
                \begin{tabular}{c@{\hspace{1ex}}l}
                    \multicolumn{2}{l}{Legend} \\
                    \midrule
                    \swatch{NeoRed} & Neoplastic \\
                    \swatch{EpiOrg} & Epithelial \\
                    \swatch{InfGrn} & Inflammatory \\
                    \swatch{ConBlu} & Connective \\
                    \swatch{DeadYel} & Dead
                \end{tabular}}
            }
        } \\[\imagegap]
        
        \raisebox{\dimexpr 0.5\imgwidth - 0.5\height\relax}[0pt][0pt]{\rotatebox{90}{\scriptsize Prediction}} &
        \includegraphics[width=\imgwidth]{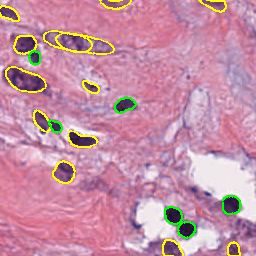} &
        \includegraphics[width=\imgwidth]{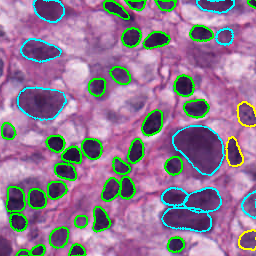} &
        \includegraphics[width=\imgwidth]{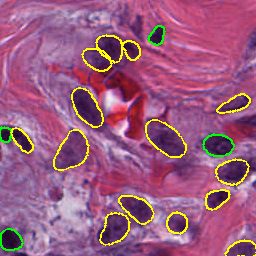} &
        \includegraphics[width=\imgwidth]{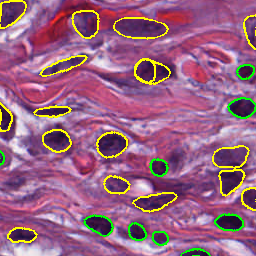} &
        \includegraphics[width=\imgwidth]{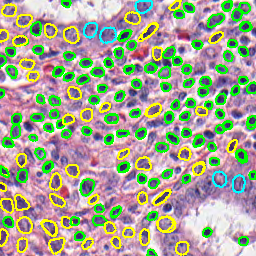} &
        \includegraphics[width=\imgwidth]{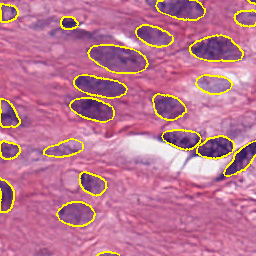} &
        \multicolumn{2}{c}{} \\
        
    \end{tabular}
    \caption{Ground-truth (GT) versus predicted segmentation masks across 19 tissue categories. Samples were randomly selected from the held-out PanNuke fold~1.}
    \label{fig:tissue_examples}
\end{figure*}

\subsection{Performance on PanNuke}

We first evaluated our method on the PanNuke dataset using standard metrics.
Because the standard Panoptic Quality (PQ) metric penalizes overlaps and PanNuke annotations are strictly non-overlapping, we applied a deterministic distance-transform watershed refinement during inference to resolve overlaps.
Under these conditions, LSP-DETR achieves competitive results (48.2 mPQ, 67.5 bPQ), placing it among top-tier methods, though slightly behind LKCell (50.3 mPQ, 68.4 bPQ).
Qualitative examples of ground-truth and predicted masks across the 19 PanNuke tissue categories are shown in \cref{fig:tissue_examples}.

However, this comparison is constrained by a fundamental limitation in the ground truth: PanNuke annotations explicitly forbid biologically realistic overlaps.
When evaluated under our proposed overlap-insensitive metrics (mMPQ, bMPQ), which remove this structural bias, LSP-DETR ranks first in binary mask quality (68.8 bMPQ) and remains highly competitive in multi-class quality (49.0 mMPQ).

A closer examination of the category-level performance reveals a notable divergence between our detection scores and segmentation quality.
While LSP-DETR achieves the highest detection F$_{1}$-score in three of the five PanNuke cell categories (\cref{tab:pannuke:classification:detection}), this does not translate to the highest category-specific PQ (\cref{tab:pannuke:classification}).
This difference is explained by the distinct aggregation methods used for each metric and the watershed post-processing required to satisfy the non-overlapping constraint.

The F$_{1}$-score is micro-averaged: true-positive (TP), false-positive (FP), and false-negative (FN) detections from the entire test set are aggregated into global sums before the F$_{1}$-score is computed.
This approach weights the final score by the performance on each individual nucleus, meaning that high-density images contribute more to the final score.

By contrast, the per-category PQ is macro-averaged: the metric is calculated independently for each image, and the final result is the arithmetic mean of these per-image scores.
This method gives equal weight to every image, regardless of the number of nuclei it contains, placing a greater emphasis on consistent performance across the entire dataset, including sparse or low-density samples.

This behavior is consistent with the training objective: the DETR-style loss function is normalized by the total number of instances within each mini-batch, mirroring the micro-averaging principle by naturally prioritizing high-density samples.
The metric divergence, therefore, reflects a real optimization bias in current DETR-style training procedures.

Future work should explore image-balanced or density-aware loss normalization so that sparse and dense samples contribute more equally during training. Evaluating such objectives with both micro- and macro-averaged metrics would clarify whether the observed category-level discrepancy can be reduced without sacrificing performance in densely populated regions.

\vspace{1.7ex}
\subsection{Generalization to MoNuSeg}
\begin{table}[!t]
    \centering
    \small
    \caption{%
        Generalization performance of models pretrained on PanNuke and evaluated on the MoNuSeg test set.
        Mean values across the three models from three-fold cross-validation are reported.
    }
    \begin{tabular*}{\linewidth}{@{}l@{\extracolsep{\fill}}cccc@{}}
        \toprule
        Model & bPQ & P & R & $\text{F}_1$  \\
        \midrule
        CellViT            & \bf 67.2 & \bf 84.7 & \bf 88.5 & \bf 86.5 \\
        LKCell                 & - & - & - & 83.0 \\
        LSP-DETR & 66.7 & 84.4 & 85.6 & 85.0 \\
        \bottomrule
    \end{tabular*}
    \label{tab:monuseg}
\end{table}

A key feature of LSP-DETR is its input-size agnosticism (\cref{tab:comparison}): despite training on small 256\texttimes 256 px patches, the model processes full-sized 1000\texttimes 1000 px MoNuSeg images in a single forward pass.

This evaluation highlights the generalizability of our model.
While LKCell achieved the highest performance on the PanNuke benchmark (\cref{tab:pannuke}), its performance drops when transferring to MoNuSeg, where it achieves an F$_{1}$-score of 83.0.
In contrast, LSP-DETR demonstrates superior robustness, outperforming LKCell with an F$_{1}$-score of 85.0 and maintaining performance comparable to the much larger CellViT model (F$_{1}$-score of 86.5).

Nevertheless, evaluation on a single external dataset provides limited evidence of domain robustness. Broader validation should include multi-institutional cohorts with different scanners, staining protocols, resolutions, organs, and disease states.

\subsection{Computational Efficiency and Scalability}

\begin{table}[bt]
    \centering
    \small
    \caption{%
        Comparison of the models in terms of computational complexity.
        *\,FLOPs were measured using a full attention layer, as the PyTorch flex attention implementation lacks a FLOPs counter.
        Here, $n$ denotes the input size (e.g., the number of pixels), and
        $k$ denotes the number of predicted nuclei instances.
    }
    \begin{tabular*}{\linewidth}{@{}l@{\extracolsep{\fill}}rrc@{}}
        \toprule
          Method      & \# Params & FLOPs & Complexity \\
          \midrule
          StarDist    &  \textbf{32.3M} &          35G\phantom{*} & $\mathcal{O}(n + k^2)$        \\
          CPP-Net     &           34.7M &          30G\phantom{*} & $\mathcal{O}(n + k^2)$        \\
          HoVer-NeXt  &           36.1M & \textbf{24G}\phantom{*} & $\mathcal{O}(n \log(n) + kn)$ \\
          LKCell      &          163.8M &         48G\phantom{*} & $\mathcal{O}(n \log(n) + kn)$ \\
          CellViT     &           699.7M &          214G\phantom{*} & $\mathcal{O}(n^2)$            \\
          LSP-DETR    &          45.0M  &                    26G* & $\bm{\mathcal{O}(n)}$         \\
          \bottomrule
    \end{tabular*}
    \label{tab:complexity}
    \vspace{-2ex}
\end{table}

The remarkable efficiency of our method is supported by a lightweight architecture (\cref{tab:complexity}): LSP-DETR has only 45M parameters and 26G FLOPs, substantially smaller than LKCell (163.8M) and CellViT (699.7M).
Furthermore, our method achieves linear complexity and generalization to larger image resolutions, properties that most transformer-based architectures (e.g., CellViT) lack due to quadratic attention cost and absolute positional encodings.
While convolutional methods can process arbitrary image sizes, their reliance on computationally expensive post-processing algorithms (with $\mathcal{O} (n \log n)$ complexity at best) makes them impractical for processing large WSI tiles efficiently.
These properties allow inference on images with resolutions of up to 8192\texttimes 8192 on a single NVIDIA H100 NVL GPU, making the method practical for high-throughput WSI pipelines.

\subsection{Ablation on the Number of Radial Rays}
\begin{table*}[b]
    \centering
    \small
    \caption{%
        Ablation study on the number of radial rays.
        All models were trained on PanNuke Fold~1 and evaluated on Fold~3 under identical settings and a fixed seed. Training time is reported for the full 130-epoch training schedule; peak GPU memory is measured during the inference benchmark.
        Ground-truth (GT) lookup memory is computed for the two float32 lookup tables of size $2 \times \text{rays} \times 256 \times 256$ used for PanNuke patches.
    }
    \begin{tabular}{lcccccccc}
        \toprule
        \multirow{2}{*}{Rays}
        & \multirow{2}{*}{mPQ} & \multirow{2}{*}{bPQ} & \multirow{2}{*}{mMPQ} & \multirow{2}{*}{bMPQ} & Inference speed & Peak GPU memory & GT lookup memory & Training time \\
            &     &     &      &      & (s/mm\textsuperscript{2}) & (GB) & (MiB/sample) & (h) \\
        \midrule
        16  & \textbf{48.3} & 66.1 & \textbf{49.1} & 67.3 & 0.45 & 58.7 & 8 & 9.8 \\
        32  & 47.8 & \textbf{66.3} & 48.6 & \textbf{67.5} & 0.45 & 58.7 & 16 & \textbf{9.7} \\
        64  & 48.1 & 66.1 & 48.9 & 67.4 & 0.45 & 58.7 & 32 & 11.2 \\
        128 & 47.8 & 65.9 & 48.7 & 67.2 & 0.45 & 57.7 & 64 & 18.2 \\
        \bottomrule
    \end{tabular}
    \label{tab:ablations:rays}
\end{table*}
\begin{table*}
    \centering
    \small
    \caption{%
        Systematic ablation study of the attention, positional-encoding, loss, matching, and decoder-depth design choices.
        Each non-baseline row changes exactly one factor from the LSP-DETR baseline, which uses local STA, RoPE, interval-based radial supervision, the inner mask cost, and $L=6$ decoder layers.
        All models were trained on PanNuke Fold~1 and evaluated on Fold~3 under identical settings and a fixed seed.
        \looseness=-1
    }
    \vspace{-1ex}
    \newcolumntype{D}{>{\scriptsize(}r<{)}}
    \begin{tabular}{lrDrDrDrD}
        \toprule
        Variant & \multicolumn{2}{c}{mPQ} & \multicolumn{2}{c}{bPQ} & \multicolumn{2}{c}{mMPQ} & \multicolumn{2}{c}{bMPQ} \\
        \midrule
        LSP-DETR (baseline) & 48.1 & \multicolumn{1}{c}{} & 66.1 & \multicolumn{1}{c}{} & 48.9 & \multicolumn{1}{c}{} & 67.4 & \multicolumn{1}{c}{} \\
        \addlinespace
        Full global attention & 48.0 & -0.1 & 65.7 & -0.4 & 48.9 & 0.0 & 67.1 & -0.3 \\
        Without RoPE & 0.0 & -48.1 & 0.0 & -66.1 & 0.0 & -48.9 & 0.0 & -67.4 \\
        Fixed-boundary radial supervision & 47.7 & -0.4 & 65.8 & -0.3 & 48.5 & -0.5 & 67.0 & -0.4 \\
        Without inner mask cost & 43.0 & -5.1 & 60.0 & -6.1 & 47.4 & -1.5 & 63.8 & -3.6 \\
        \addlinespace
        Decoder depth $L=3$ & 45.7 & -2.4 & 64.6 & -1.5 & 46.8 & -2.1 & 66.2 & -1.2 \\
        Decoder depth $L=9$ & 48.7 & +0.6 & 66.1 & +0.0 & 49.5 & +0.6 & 67.3 & -0.1 \\
        \bottomrule
    \end{tabular}
    \label{tab:ablations:components}
\end{table*}

Because the star-convex shape descriptor directly determines the angular sampling of each predicted contour, the number of radial rays controls a trade-off between contour resolution and computational cost.
Starting from our initially selected 64-ray configuration, we trained additional models with 16, 32, and 128 rays using the same training protocol.
All models were trained on PanNuke Fold~1 and evaluated on Fold~3 to isolate the effect of the ray count under a fixed data split.

The results in \cref{tab:ablations:rays} show that increasing the number of rays beyond 32 does not improve segmentation accuracy in this experiment. The 32-ray model achieves the highest bPQ (66.3) and bMPQ (67.5), while its mPQ and mMPQ remain within 0.5 percentage points of the best values observed across the ablation. Inference speed and peak GPU memory are effectively unchanged across the evaluated ray counts, indicating that the radial descriptor has a negligible effect on GPU memory usage in practice. Compared with the 64-ray configuration, the 32-ray configuration halves the ground-truth lookup memory (16 versus 32~MiB per sample) and reduces training time from 11.2~h to 9.7~h. Conversely, the 128-ray configuration increases training time to 18.2~h without improving any reported accuracy metric.

Thus, for this experimental setting, 32 rays provide the most favorable accuracy--cost trade-off and would be a reasonable default choice. We retain 64 rays for the main experiments because this configuration was selected \emph{a priori}, following prior StarDist work~\cite{StarDistClassification}, before conducting the present ablation. The comparable performance at 32 rays suggests that future deployments can use the lower-dimensional descriptor when reductions in training time and supervision-memory requirements are important.

\vspace{1ex}
\subsection{Architecture Ablations}

We systematically isolate the contributions of the main architectural and loss-design choices in \cref{tab:ablations:components}.
Every non-baseline variant changes one component while all remaining components, the training protocol, seed, and the Fold~1 to Fold~3 evaluation split are held fixed.
Grid-query initialization is retained throughout to preserve the input-size-adaptive query layout of the proposed model.

Full global attention did not improve segmentation accuracy over local STA.
STA is therefore preferred because it retains the accuracy of full attention while changing the attention complexity from quadratic to linear in the input size.
RoPE is essential for optimization: without it, the model fails to learn, yielding zero scores for all segmentation metrics.
The inner mask cost in bipartite matching is also important, as removing it causes a substantial drop in all metrics.
Increasing the decoder depth from six to nine layers yields only marginal gains in mPQ and mMPQ, no change in bPQ, and a slight decrease in bMPQ. Because this deeper configuration also increases FLOPs from 26G to 33G, the six-layer decoder offers a more efficient accuracy--complexity trade-off.

\subsection{Biologically Realistic Overlap Modeling}

\begin{table}[t]
    \centering
    \small
    \caption{Expert consensus annotations, sample sizes, and median model-predicted intersection areas for the evaluated pairs of nuclei.}
    \label{tab:agreement_areas}
    \begin{tabular*}{\linewidth}{@{}l@{\extracolsep{\fill}}cc@{}}
        \toprule
        Consensus Annotation & $n$ & Median Predicted Intersection \\
        \midrule
        Unsure      & 24 & $3.1 \text{ \textmu m}^2$ \\
        Overlapping & 78 & $2.9 \text{ \textmu m}^2$ \\
        Touching    & 37 & $0.0 \text{ \textmu m}^2$ \\
        \bottomrule
    \end{tabular*}
    \vspace{-2ex}
\end{table}

To our knowledge, LSP-DETR is the first method that enables biologically plausible modeling of overlaps between nuclei without requiring explicit overlap annotations during training. This capability is a significant advantage given the prevalence of overlapping cell structures in real tissue.
A major limitation in computational pathology is the lack of datasets with precise, pixel-level annotations for overlapping instances; standard ground-truth masks are artificially separated.
Consequently, it is impossible to quantitatively evaluate overlap segmentation with standard metrics.

\begin{figure}[!t]
    \centering
    \begin{subfigure}[b]{0.49\linewidth}
        \centering
        \includegraphics[width=\linewidth]{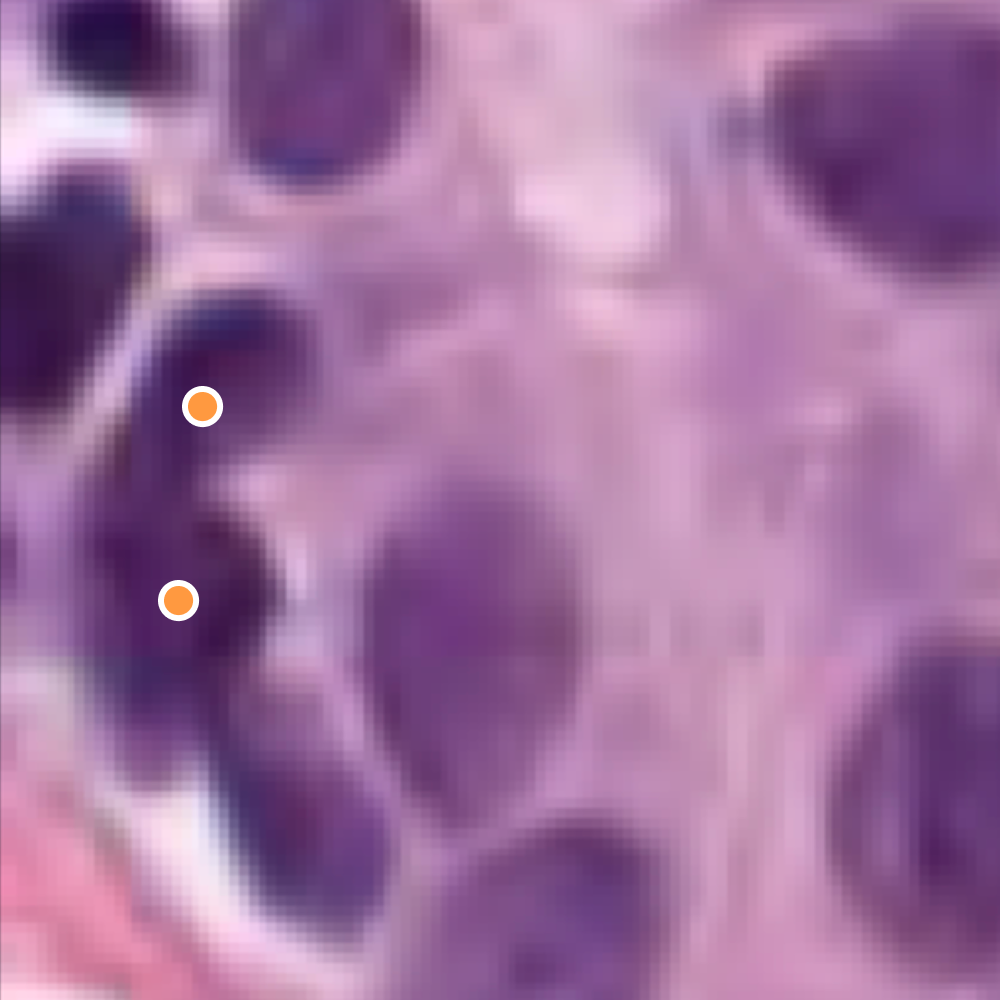}
    \end{subfigure}
    \hfill
    \begin{subfigure}[b]{0.49\linewidth}
        \centering
        \includegraphics[width=\linewidth]{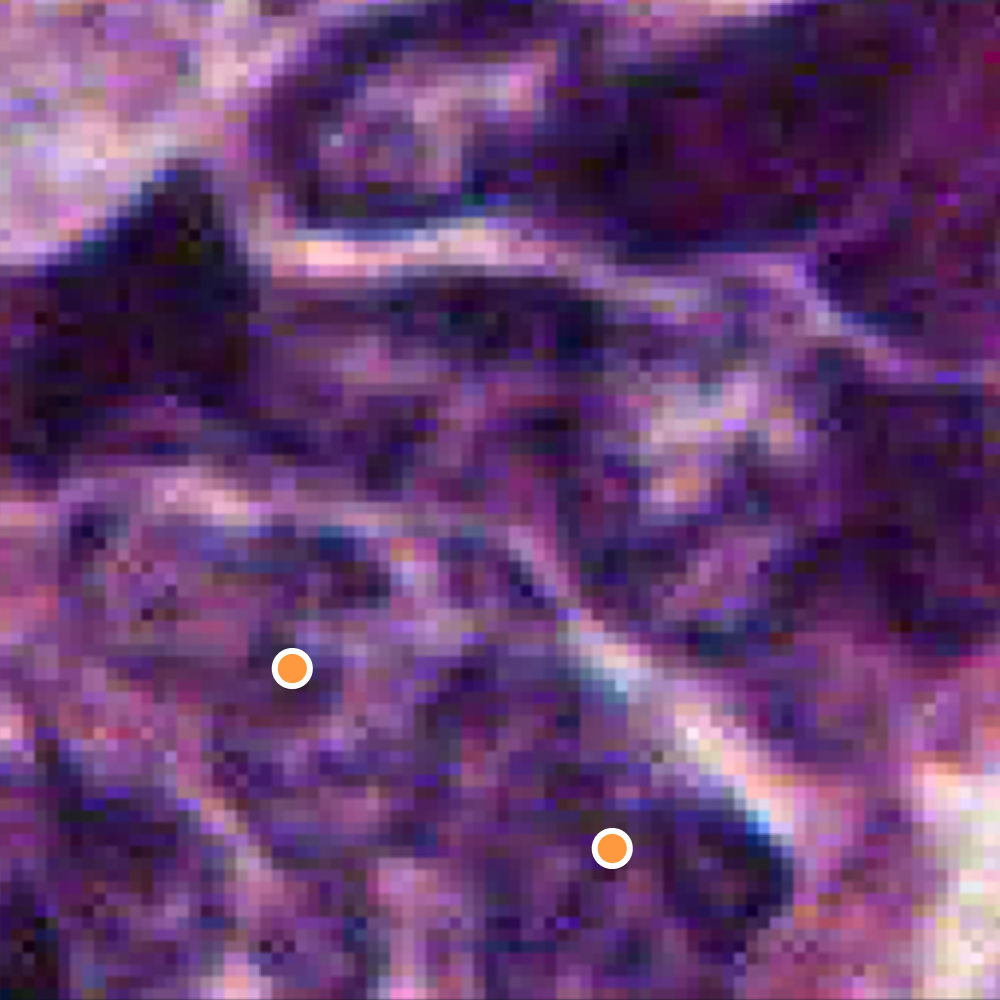}
    \end{subfigure}
    \\
    \vspace{0.02\linewidth}
    \begin{subfigure}[b]{0.49\linewidth}
        \centering
        \includegraphics[width=\linewidth]{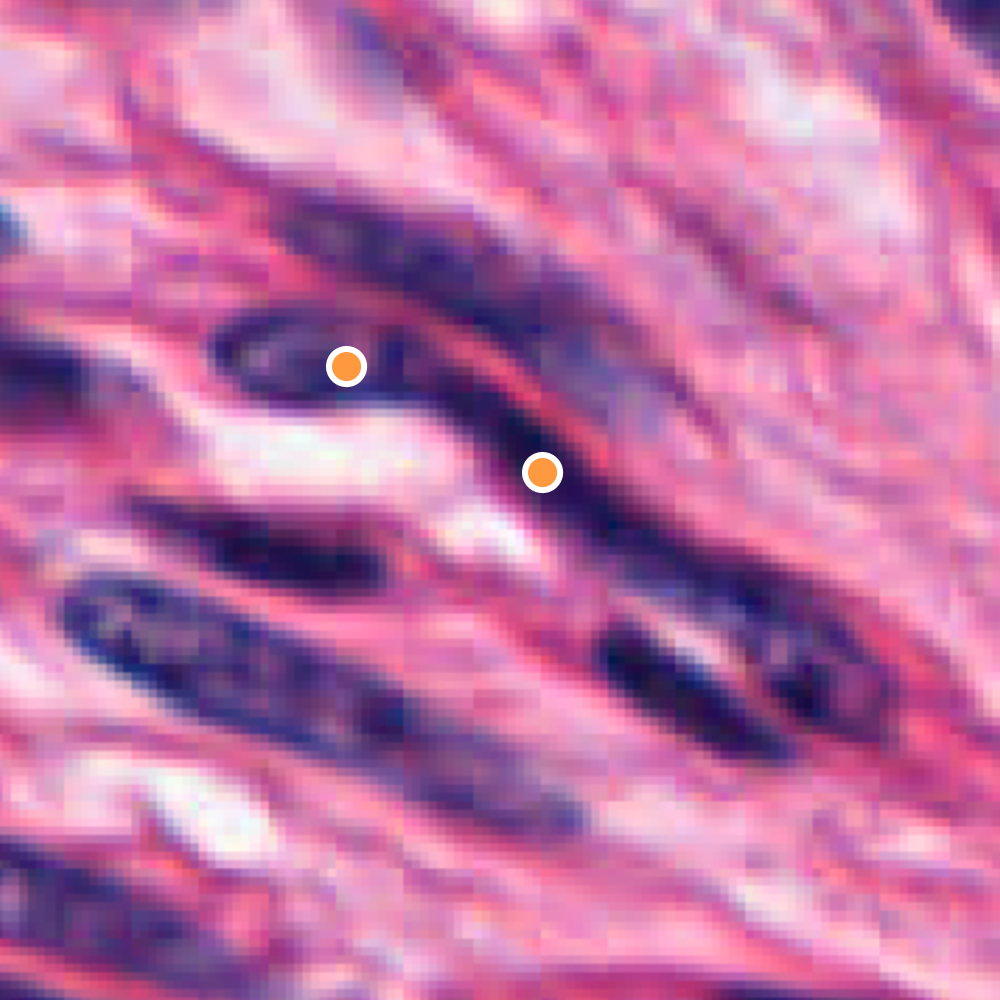}
    \end{subfigure}
    \hfill
    \begin{subfigure}[b]{0.49\linewidth}
        \centering
        \includegraphics[width=\linewidth]{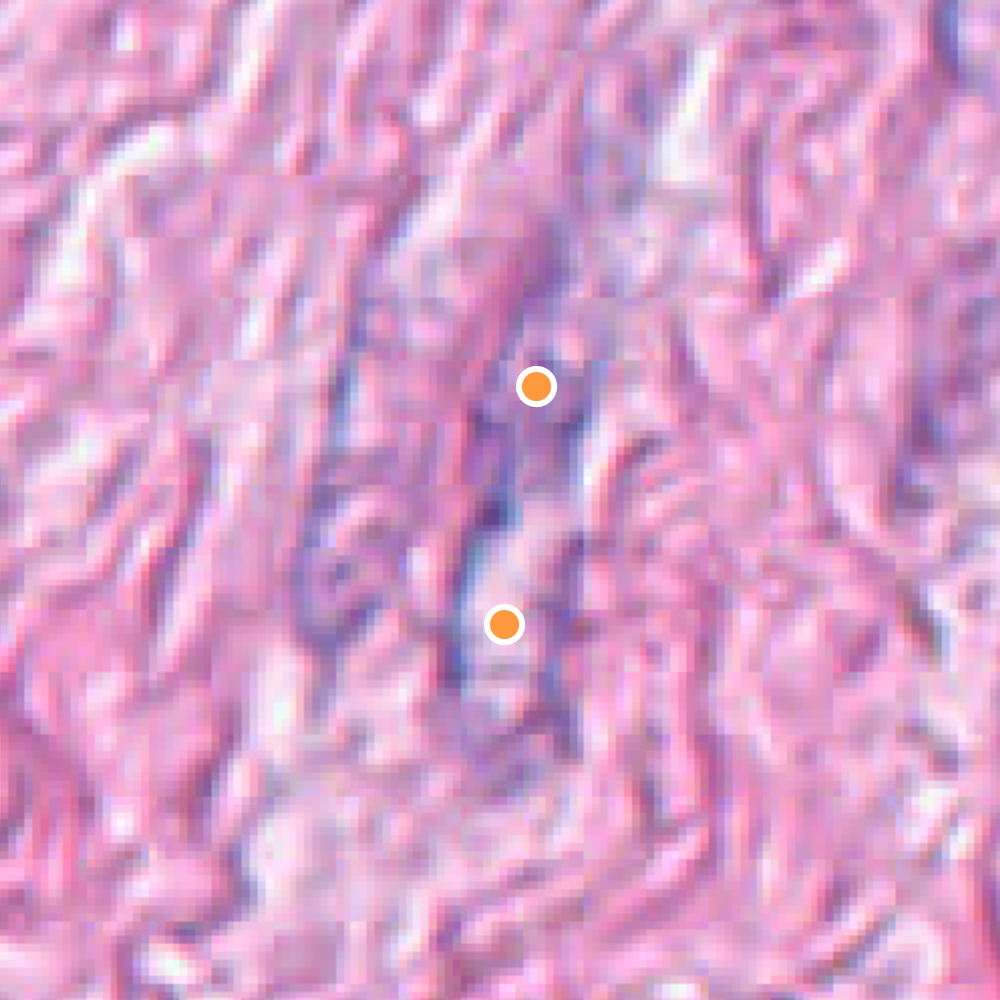}
    \end{subfigure}
    \caption{Examples for which both pathologists agreed that it was impossible to determine whether each pair of nuclei was touching or overlapping. Orange dots represent the model-predicted centroids of the evaluated pairs.}
    \label{fig:undecidable_cases}
    \vspace{4ex}
\end{figure}

\begin{figure}[!t]
    \centering
    \begin{subfigure}{\linewidth}
        \centering
        \includegraphics[width=\linewidth]{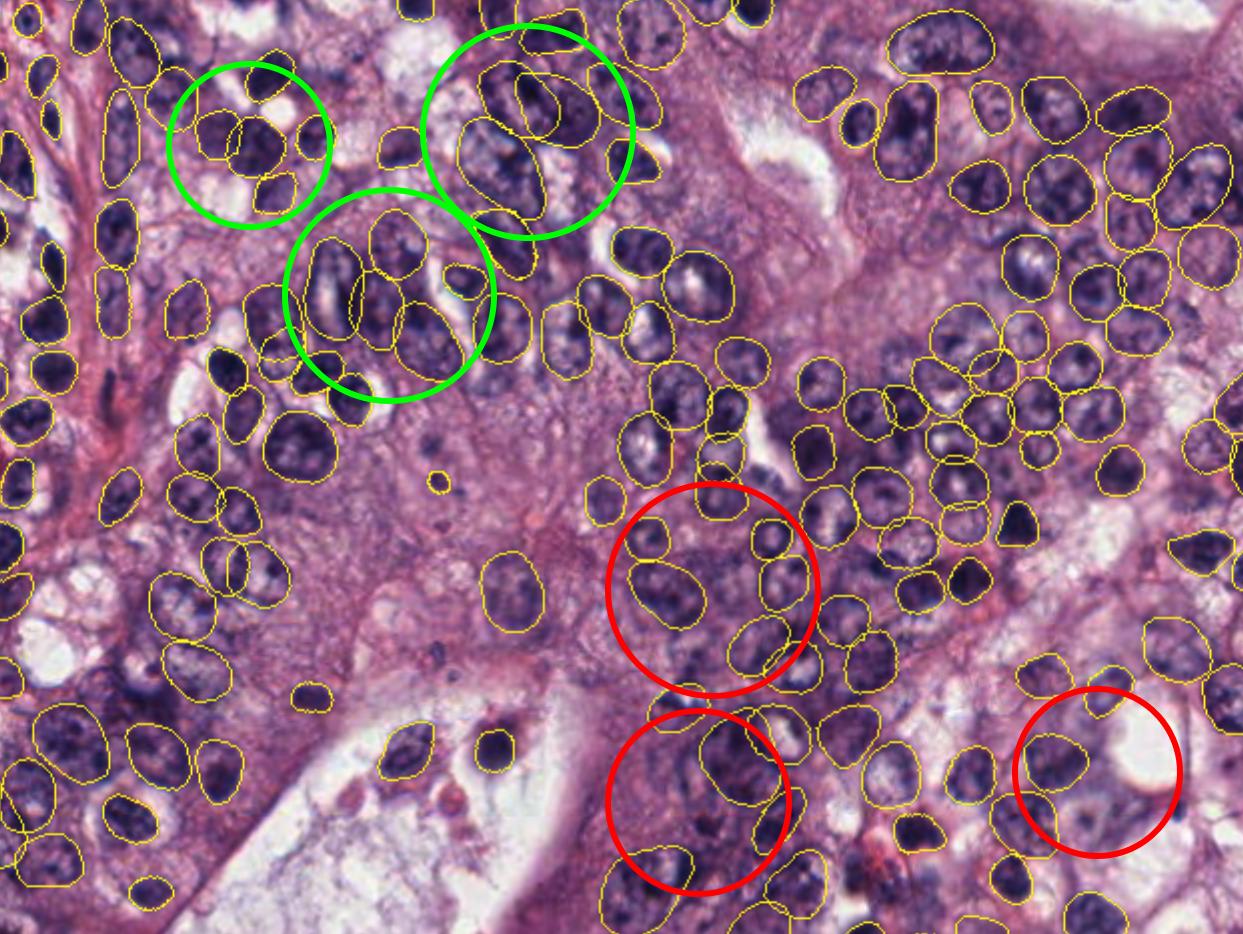}
    \end{subfigure}
    \\
    \vspace{0.02\linewidth}
    \begin{subfigure}{0.49\linewidth}
        \centering
        \includegraphics[width=\linewidth]{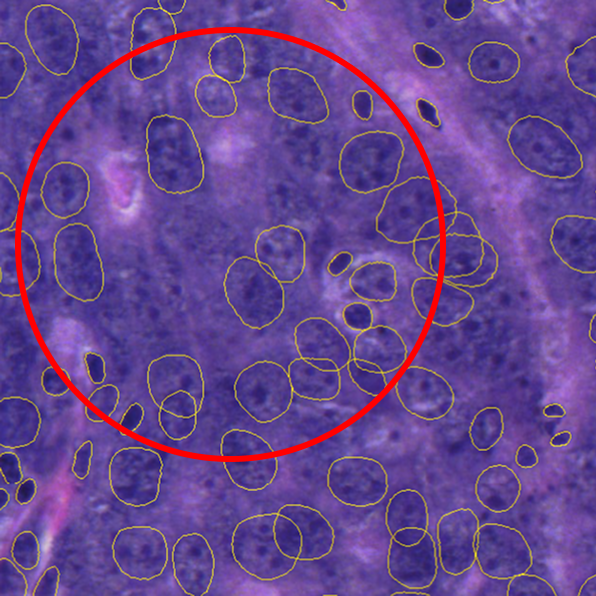}
    \end{subfigure}
    \hfill
    \begin{subfigure}{0.49\linewidth}
        \centering
        \includegraphics[width=\linewidth]{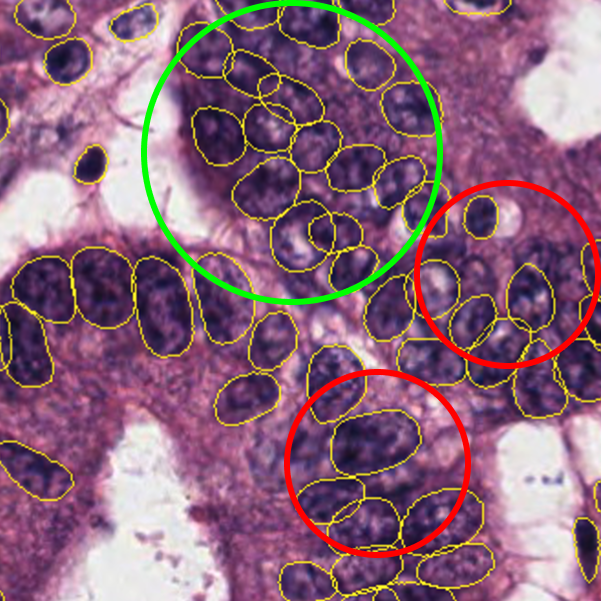}
    \end{subfigure}
    \caption{Model performance on H\&E-stained colon cancer slides of average quality. The model can identify overlapping nuclei to some extent (green labels) but fails in areas with poor focus or contrast and poorly discernible nuclear boundaries (red labels).}
    \label{fig:overlaps-example}
\end{figure}

We therefore complemented the standard benchmark evaluation with a targeted pathologist review focusing on the clinically relevant distinction between mere touching and genuine overlap in neighboring nuclei. The resulting pooled F$_1$-score of 0.964 and AUROC of 0.983 demonstrate that the magnitude of LSP-DETR's predicted polygon intersections aligns strongly with expert consensus. Notably, our analysis of ambiguous pairs (\cref{tab:agreement_areas}) revealed a counterintuitive trend: the median predicted area for pairs classified by consensus as \say{unsure} ($3.1 \text{ \textmu m}^2$) exceeded that of pairs confidently classified as \say{overlapping} ($2.9 \text{ \textmu m}^2$). Visual inspection of these undecidable pairs (\cref{fig:undecidable_cases}) confirms that this uncertainty is not driven by microscopic, near-zero intersections. Rather, it is caused by low resolution and heavily blurred borders of the nuclei, where true cellular boundaries are entirely obscured. In these instances, the model overestimates the intersection area, precisely mirroring the visual ambiguity that prevents experts from making a definitive call. Overall, this targeted validation supports the qualitative observation (\cref{fig:overlaps-example}) that LSP-DETR effectively models biologically realistic overlaps, successfully delineating instances that traditional methods would typically merge or artificially split.
\looseness=-1

\paragraph{Ablation on Overlap Modeling}

This ablation explains the fixed-boundary row in \cref{tab:ablations:components}.
It compares our interval-based radial-distance loss with the conventional single-boundary formulation, which typically penalizes any instance overlap during training.
We simulate the latter by setting $r_{\max} = r_{\min}$, thereby forcing the model to converge on one precise boundary rather than a plausible radial interval.
The fixed-boundary ablation consistently decreases accuracy, supporting the benefit of interval-based radial supervision.

\subsection{Potential Clinical Relevance, Limitations, and Future Work}
\label{sec:clinical-relevance}

Low on-demand latency is most relevant when a pathologist delineates or revises an H\&E region for molecular testing during slide review, rather than when slides are processed in an offline batch. A direct clinical application is pathologist-supervised estimation of neoplastic cellularity (histological tumor-cell fraction), defined using the numbers of neoplastic and non-neoplastic nuclei within a selected region of viable tissue. Visual estimates are affected by inconsistent use of cell-count-based and area-based definitions and show substantial interobserver variability, particularly in specimens with dense inflammatory infiltrates \cite{lhermitte2017adequately,devereaux2022neoplastic}. Because tumor, stromal, and inflammatory nuclei differ systematically in size, morphology, and local density, merge errors among small, densely packed lymphocyte nuclei could preferentially undercount the non-neoplastic population, overestimate neoplastic cellularity, and consequently underestimate the expected admixture of non-neoplastic DNA. After task-specific training and validation, a pathologist-reviewable overlay of individually segmented and classified nuclei could provide auditable counts and rapid recalculation after the extraction region is adjusted; computer-assisted H\&E cell counting has been shown to improve agreement with reference counts and reduce interobserver variability \cite{frei2023pathologist,limperio2025digital}. For large or repeatedly revised regions, the measured per-area speed advantage of LSP-DETR (\cref{tab:comparison,tab:wsi}) could therefore shorten on-demand analysis during the clinical interaction.

This represents a potential deployment setting rather than a clinically validated application of the present model. The expert overlap experiment establishes an association between predicted polygon intersection and pathologist assessment of overlap versus contact; it does not establish accurate neoplastic-versus-non-neoplastic classification, improved tumor-cell-fraction assessment or molecular-test interpretation, recovery of the unobserved complete contour of a nucleus, or improvement in histological grading. We therefore interpret the predictions as plausible overlap-aware instance geometries rather than reconstructions of the true three-dimensional shape of a nucleus.

Future work should address these limitations through several complementary directions. First, overlap modeling should be evaluated on purpose-built datasets containing independently annotated visible and occluded nuclear boundaries, ideally supplemented by serial sections, confocal microscopy, or three-dimensional imaging where the underlying nuclear geometry can be observed more directly. Such datasets would permit quantitative evaluation of both overlap detection and contour completion instead of relying only on pairwise expert classification. Second, the model should be externally validated across institutions, scanners, staining protocols, tissue types, and pathological conditions. This evaluation should include prospective specimens and should separately quantify performance in difficult regions characterized by blur, artifacts, necrosis, inflammation, or exceptionally dense nuclei. Domain-robust training, stain augmentation, and targeted fine-tuning on underrepresented conditions could then be used to address the identified failure modes.

Third, the star-convex representation may be insufficient for strongly concave, fragmented, or highly irregular nuclei. Future variants could investigate more flexible representations, such as Gaussian splats, splines, or Fourier descriptors, while retaining direct set prediction and avoiding expensive post-processing. The discrepancy between micro-averaged detection performance and macro-averaged PQ also indicates that the current instance-normalized objective favors dense images. This could be addressed by introducing image-balanced or density-aware loss weighting.
\looseness=-1

Finally, clinical utility must be established through task-specific studies rather than inferred from segmentation accuracy alone. For neoplastic-cellularity estimation, the model should be evaluated prospectively against expert counts, molecular-quality outcomes, and interobserver variability. A reader study could compare unaided assessment with pathologist-supervised use of editable LSP-DETR overlays, measuring accuracy, review time, correction burden, and failure detectability. Similar task-specific validation would be required before claiming benefits for grading or other diagnostic applications. These studies would determine whether the model’s computational efficiency and overlap-aware predictions translate into reliable improvements in clinical practice.

\section{Conclusions}

In this work, we introduced LSP-DETR, a novel and fully end-to-end framework for nuclei instance segmentation.
By reimagining nuclei segmentation as a direct set prediction task and representing nuclei as star-convex polygons, we have eliminated the need for computationally expensive, handcrafted post-processing steps that have long hindered the scalability of digital pathology workflows.

Our experimental results on the PanNuke, MoNuSeg, and TCGA datasets demonstrate that LSP-DETR is a highly effective and efficient solution:

\paragraph{State-of-the-Art Efficiency}
With an inference speed of 0.45 s/mm\textsuperscript{2} and linear computational complexity, our model is more than three times faster than the next-fastest leading method.

\paragraph{Biologically Realistic Modeling}
Through our novel radial distance loss, the model naturally learns to segment overlapping nuclei---a common occurrence in real tissue---without requiring explicit overlap annotations during training.
    
\paragraph{Input-Size Agnosticism}
The model demonstrates input-size scalability, successfully processing large-scale images in a single forward pass despite being trained on smaller patches.

\bigskip
In conclusion, LSP-DETR bridges the gap between high-fidelity segmentation and clinical requirements, setting a new direction in nuclei segmentation through direct instance modeling.

\section*{Compliance with Ethical Standards}

\paragraph{Ethical Approval}

The project was approved by the Ethical Committee of Masaryk Memorial Cancer Institute, No.~MOU 385\,920.

\paragraph{Acknowledgements}

This work has been supported by the AI infrastructure developed in the BioMedAI TWINNING project, supported by the EU Horizon Europe program under grant agreement no.~101079183.
CERIT-SC computational resources were provided by the e-INFRA CZ project (ID:90140), supported by the Ministry of Education, Youth, and Sports of the Czech Republic.

\paragraph{Declaration of generative AI and AI-assisted technologies in the manuscript preparation process}

During the preparation of this work, the authors used ChatGPT and Gemini to improve the formulation and fluency of the text. Additionally, GitHub Copilot was used to assist with code development and optimization. After using these tools, the authors reviewed and edited the content as needed and take full responsibility for the content of the published article.

\end{document}